\documentclass[10pt,twocolumn,journal]{IEEEtran}

\usepackage{cite}
\usepackage{amsmath,amssymb,amsfonts}
\usepackage{algorithm}
\usepackage{algpseudocode}
\usepackage{graphicx}
\usepackage{textcomp}
\usepackage{booktabs}
\usepackage{multirow}
\usepackage{threeparttable}
\usepackage{array}
\usepackage{makecell}
\usepackage{placeins}
\usepackage{url}
\usepackage{hyperref}
\hypersetup{hidelinks}

\algrenewcommand\algorithmicrequire{\textbf{Require:}}
\algrenewcommand\algorithmicensure{\textbf{Ensure:}}
\makeatletter
\renewcommand{\ALG@name}{Algorithm}
\makeatother

\newcommand{\argmin}{\operatorname*{arg\,min}}

\def\BibTeX{{\rm B\kern-.05em{\sc i\kern-.025em b}\kern-.08em
    T\kern-.1667em\lower.7ex\hbox{E}\kern-.125emX}}

\providecommand{\IEEEmembership}[1]{#1}

\begin{document}

\title{Diffusion-Encoding Gaussian Field for Joint k--q dMRI Reconstruction}

\author{
Zhibo~Chen,
Yajuan~Huang,
Yu~Guan,
Qiuyun~Fan,
\\Dong~Liang,~\IEEEmembership{Senior Member,~IEEE},
and Qiegen~Liu,~\IEEEmembership{Senior Member,~IEEE}%
\thanks{This work was supported in part by the National Key Research and Development Program of China under Grant 2023YFF1204300 and Grant 2023YFF1204302. (Corresponding author: Qiegen Liu).}
\thanks{Zhibo Chen, Yajuan Huang, and Qiegen Liu are with the School of Information Engineering, Nanchang University, Nanchang 330031, China (e-mail: 41610024022@email.ncu.edu.cn; 6105123171@email.ncu.edu.cn; liuqiegen@ncu.edu.cn).}
\thanks{Yu Guan is with the School of Advanced Manufacturing and the School of Information Engineering, Nanchang University, Nanchang 330031, China (e-mail: guanyu@ncu.edu.cn).}
\thanks{Qiuyun Fan is with the Academy of Medical Engineering and Translational Medicine, Medical School, Faculty of Medicine, Tianjin University, Tianjin, 300072, China (e-mail: fanqiuyun@tju.edu.cn).}
\thanks{Dong Liang is with the Lauterbur Research Center for Biomedical Imaging and the Research Center for Medical AI, Shenzhen Institute of Advanced Technology, Chinese Academy of Sciences, Shenzhen 518055, China (e-mail: dong.liang@siat.ac.cn).}
}

\maketitle

\begin{abstract}
\end{abstract}

\begin{IEEEkeywords}
Diffusion MRI, Gaussian splatting, self-supervised reconstruction, spatial--angular representation.
\end{IEEEkeywords}

\begin{figure}[!t]
\centering
\includegraphics[width=\columnwidth]
{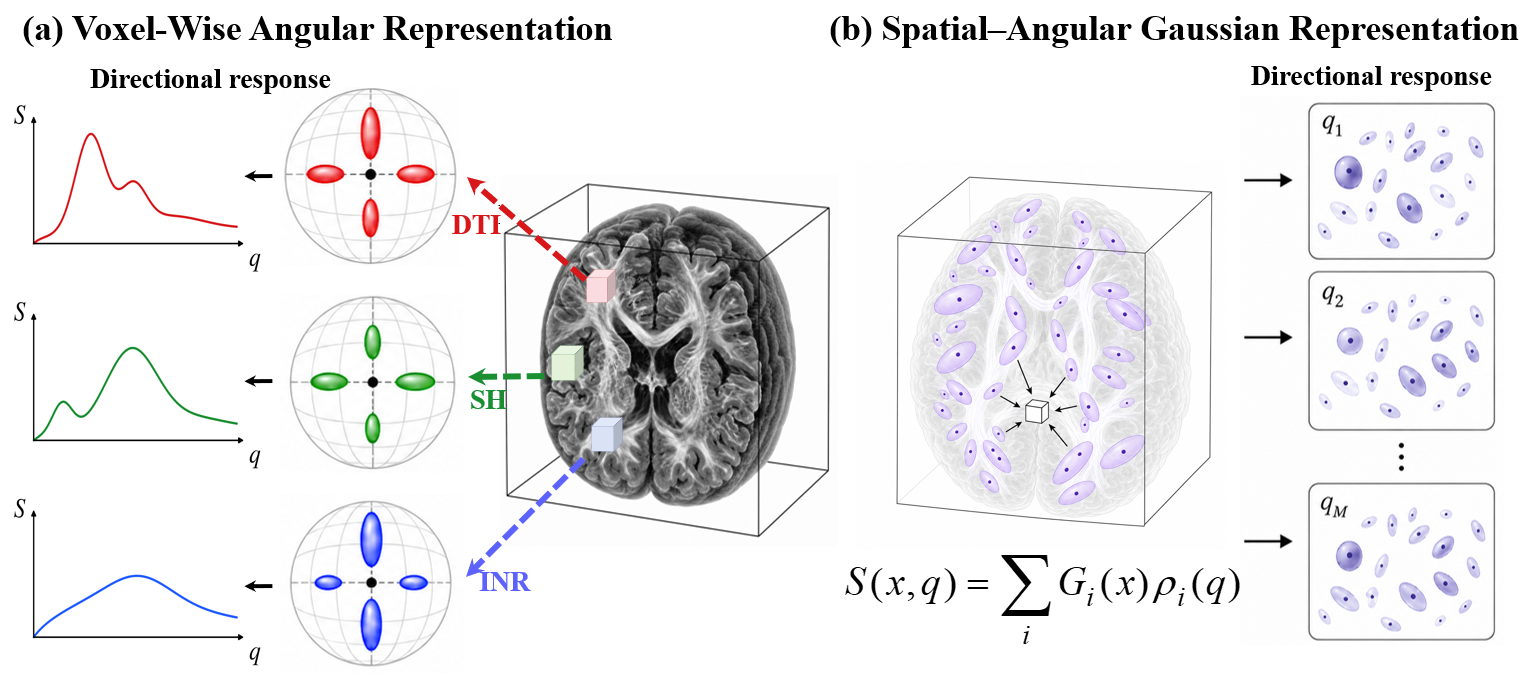}
\caption{
Comparison of voxel-centered and primitive-centered spatial--angular
dMRI representations.
(a) A directional response is independently attached to each fixed voxel,
 (b) overlapping 3D Gaussian primitives provide shared spatial support
and q-conditioned responses whose weighted aggregation yields the queried
DWI signal.
}
\label{fig:overview}
\end{figure}

\section{Introduction}
\label{sec:introduction}

\IEEEPARstart{D}{iffusion} magnetic resonance imaging (dMRI) enables noninvasive
characterization of tissue microstructure and white-matter organization~\cite{r_stejskal1965,r_basser1994_dti,r_lebihan2001}.
It requires repeated image acquisitions
under different diffusion-encoding directions.
Consequently, the acquisition time increases with both the spatial sampling
density in k-space and the angular sampling density in q-space.
Undersampling k-space shortens each diffusion-weighted acquisition,
whereas reducing q-space sampling decreases the number of repeated
acquisitions.
However, these two forms of undersampling are not independent.
Errors introduced by k-space undersampling alter the local signal
intensities from which angular responses are estimated, and may therefore
be misinterpreted as genuine directional variation when q-space is sparse.
The resulting errors can propagate to tensor-derived metrics and orientation
estimates, including fractional anisotropy (FA), mean diffusivity (MD), and
principal diffusion directions~\cite{r_basser1996_metrics}.
Joint k--q reconstruction must therefore preserve not only image-domain
spatial fidelity, but also the directional relationships required for reliable
diffusion quantification.

Existing joint k--q reconstruction methods exploit joint spatial--angular
sparsity or incorporate learned q-space priors into measurement-consistent
optimization frameworks~\cite{r_kqcs2017,r_qmodel2020}.
More broadly, model-based and self-supervised MRI reconstruction methods
combine learned priors with data consistency to reduce their dependence on
fully sampled training targets~\cite{r_modl2019,r_ssdu2020}.
Learning-based approaches have also addressed q-space acceleration through
direct regression, multimodal or attention-based estimation, recurrent
architectures, spatio-angular convolutions, and generative angular priors
~\cite{r_qdl2016,r_msrqdl2021,r_aqdl2025,r_rcnn2022,r_pccnn2023,r_pgdit2025}.

Despite these advances, most existing formulations couple spatial and
angular information only indirectly through regularization or learned
features, or process spatial reconstruction and angular completion
sequentially.
Diffusion-weighted images acquired under different directions share the
same spatially aligned anatomy, while their local signal intensities vary
with diffusion encoding.
Measurements across directions should therefore jointly constrain a common
spatial representation, while direction-dependent responses account for
genuine diffusion-related signal variation.
However, indirect or sequential formulations lack an explicit shared spatial
support for integrating complementary measurements across directions during
undersampled reconstruction.

Continuous neural fields support representing and querying signals at
unobserved directions, but their local spatial support is generally encoded
implicitly in network weights or coefficient fields
~\cite{r_siren2020,r_nesh2023,r_sarl2025,r_sainr2026}.
In contrast, Gaussian primitives provide explicit, learnable, and spatially
localized support
~\cite{r_3dgs2023,r_3dgsmr2025}.
A common set of Gaussian primitives can therefore serve as an explicit
anatomical scaffold shared across diffusion directions.
However, pairing a shared 3D Gaussian scaffold with a separate angular model
still leaves spatial support and directional response decoupled.
Because the angular model cannot directly adapt the underlying primitive
support, spatial fitting errors may be absorbed as apparent directional
variation and propagated to unobserved directions, as illustrated in
Fig.~\ref{fig:overview}(a).
The diffusion response should therefore be embedded into the Gaussian
primitives themselves rather than appended after spatial reconstruction.

To close this gap, we propose a self-supervised diffusion-encoding
spatial--angular Gaussian field, as illustrated in
Fig.~\ref{fig:overview}(b).
All diffusion directions share a common set of overlapping 3D Gaussian primitives, with each primitive carrying a continuous q-conditioned response. The signal at each spatial location is synthesized from multiple neighboring primitive responses.
This primitive-centered formulation couples local spatial support and
directional attenuation within the same explicit units.
Each response combines a positive-semidefinite diffusion-tensor attenuation
anchor with a regularized even angular correction, balancing physical
stability and angular flexibility.
To stabilize estimation from sparse joint k--q measurements, discrete
primitive responses are first calibrated at the observed directions, while
spatial reconstruction discrepancy, sampled k-space inconsistency, and
cross-direction angular heterogeneity jointly guide primitive birth and
splitting.
The calibrated responses are then projected onto a tensor-anchored continuous
q-response and jointly refined with the updated Gaussian support.
All stages are driven only by sampled k-space measurements from observed
directions, and the resulting field is queried for both observed-direction
reconstruction and held-out-direction synthesis.
The method is evaluated independently on three HCP diffusion shells under
multiple combinations of spatial and angular acceleration.

The main contributions of this work are summarized as follows:
\begin{itemize}

\item
We introduce an explicit dMRI representation with a Gaussian scaffold shared
across diffusion directions and a continuous q-conditioned response carried
by each primitive.
Through one-to-many primitive support and many-to-one response aggregation,
the resulting field jointly models shared anatomy and direction-dependent
attenuation.

\item
We develop a dMRI-oriented strategy that jointly adapts Gaussian support
and primitive q-responses using observed-direction measurements.
Spatial discrepancy, back-projected k-space residuals, and angular
heterogeneity guide primitive birth and splitting, after which calibrated
responses are projected onto a positive tensor-anchored continuous
q-response for self-supervised synthesis at unobserved directions.

\end{itemize}

\section{Related Work}
\label{sec:related_work}

\subsection{Joint k--q Reconstruction}
\label{sec:rw_kq}

Diffusion MRI acquisition depends on both k-space and q-space. The former
determines image spatial structure, whereas the latter describes signal
variation across diffusion directions. Insufficient k-space sampling may
produce aliasing artifacts and loss of spatial detail, while sparse
diffusion-direction sampling may affect diffusion-tensor fitting, fiber
orientation estimation, and microstructural measures such as FA and MD.
Accelerated dMRI reconstruction must therefore consider both spatial and
directional sampling.

Existing studies have extended compressed-sensing MRI~\cite{r_lustig2007} to
joint k--q reconstruction. For example, $(k,q)$-CS recovers diffusion images
by exploiting joint spatial--angular sparsity~\cite{r_kqcs2017}, whereas
qModeL performs model-based reconstruction using a learned q-space prior
within a measurement-consistent framework~\cite{r_qmodel2020}. These methods
directly handle undersampled measurements and provide an important foundation
for joint k--q dMRI reconstruction. However, most existing approaches
represent dMRI as voxel-grid DWI stacks or q-space coefficient fields, in
which shared anatomical structure across diffusion directions and local
direction-dependent responses are modeled less explicitly.

\subsection{q-Space Modeling}
\label{sec:rw_qspace}

The dMRI signal varies with the diffusion-gradient direction, making
directional relationships in q-space essential for diffusion-signal
modeling. Diffusion tensor imaging uses a tensor model to describe
direction-dependent signal attenuation and remains one of the most widely
used diffusion models~\cite{r_basser1994_dti}. Beyond DTI, Q-ball imaging and
regularized spherical-harmonic formulations characterize higher-order angular
structure, while Gaussian-process regression supports interpolation of
nonuniform or undersampled q-space measurements~\cite{r_tuch2004,
r_descoteaux2007,r_sjolund2016}.

Deep-learning methods have further advanced q-space modeling. Early q-space
deep learning directly maps sparse diffusion measurements to microstructural
quantities~\cite{r_qdl2016}, while multimodal and attention-based approaches
improve angular super-resolution or accommodate variable q-space sampling
strategies~\cite{r_msrqdl2021,r_aqdl2025}. Recurrent convolutional
autoencoders model relationships among diffusion directions~\cite{r_rcnn2022},
and spatio-angular convolutions jointly extract spatial and directional
features~\cite{r_pccnn2023}. Spatial--angular representation learning and
implicit representations model dMRI as a continuous spatial-directional
signal~\cite{r_sarl2025,r_sainr2026}, while physics-guided generative priors
have also been used for high-angular-resolution diffusion-image
synthesis~\cite{r_pgdit2025}. These studies demonstrate the value of
diffusion priors and q-space continuity for recovering complete diffusion
signals from sparse directions, and motivate direction-conditioned response
modeling.

\subsection{Gaussian Splatting}
\label{sec:rw_gaussian}

Medical image reconstruction depends not only on the reconstruction
algorithm, but also on how the image signal is represented. In addition to
conventional voxel-grid representations, implicit neural representations
such as NeRF and SIREN describe images or scenes as continuous
coordinate-conditioned functions~\cite{r_nerf2020,r_siren2020}. NeRP further
applies this principle to sparsely sampled medical image
reconstruction~\cite{r_nerp2021}. These studies show that continuous
representations can provide effective priors for medical inverse problems.

In contrast to implicit coordinate networks, 3D Gaussian Splatting represents
a three-dimensional field using explicit Gaussian primitives with learnable
centers, scales, and appearance attributes, which are optimized through
differentiable splatting~\cite{r_3dgs2023}. Gaussian representations have
subsequently been extended to dynamic scene modeling~\cite{r_4dgs2023} and
medical inverse problems, including sparse-view CT, dynamic angiography,
4D CT, and undersampled MRI reconstruction~\cite{r_r2gaussian2024,
r_4drgs2024,r_4dctgs2025,r_3dgsmr2025}.

For dMRI, different diffusion directions produce direction-dependent signal
variations under the same anatomical structure. The explicit spatial
attributes of Gaussian primitives can therefore describe shared anatomical
support, while direction-conditioned responses attached to the primitives
can model q-space signal variation. This provides the basis for jointly using
Gaussian representations for spatial reconstruction and diffusion-direction
modeling.

\section{Methodology}
\label{sec:method}

\begin{figure*}[!t]
\centering
\includegraphics[width=\textwidth]
{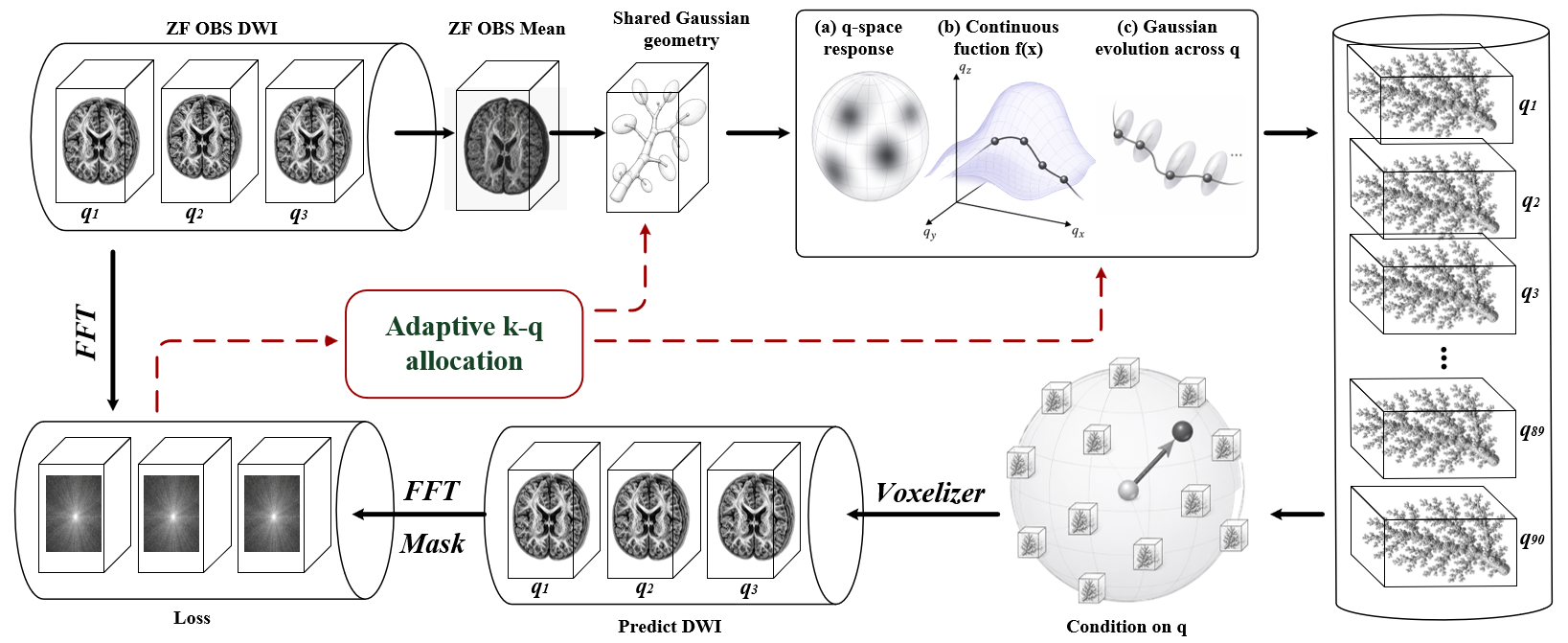}
\caption{
Overview of the proposed diffusion-encoding spatial--angular Gaussian
framework. A shared Gaussian scaffold with continuous tensor--residual
responses is optimized using observed-direction reconstruction and sampled
k-space consistency, while fitting residuals guide adaptive primitive
allocation.
}
\label{fig:method_overview}
\end{figure*}

Our method consists of two main components: a diffusion-encoding
spatial--angular Gaussian field that represents the DWI signal using shared
3D Gaussian primitives with continuous tensor-anchored q-responses
(Sec.~\ref{sec:spatial_angular_field}), and a progressive field construction
strategy that performs scaffold initialization, observed-response
calibration, adaptive primitive allocation, continuous-response projection,
and joint refinement
(Sec.~\ref{sec:progressive_optimization}).

\subsection{Diffusion-Encoding Spatial--Angular Gaussian Field}
\label{sec:spatial_angular_field}

\subsubsection{Joint k--q Inverse Problem}
\label{sec:joint_kq_problem}

We consider one fixed diffusion-weighting shell at a time.
Let $S(\mathbf{x},\mathbf{g})$ denote the diffusion-weighted signal at spatial
coordinate $\mathbf{x}\in\Omega_x\subset\mathbb{R}^{3}$ and unit
diffusion-encoding direction $\mathbf{g}\in\mathbb{S}^{2}$,
the k-space measurements from observed diffusion directions are modeled as
\begin{equation}
\mathbf{y}_{\mathbf{g}}
=
\mathbf{M}_{\mathbf{g}}
\odot
\mathcal{F}_{xy}
\left(
S(\cdot,\mathbf{g})
\right)
+
\boldsymbol{\epsilon}_{\mathbf{g}},
\qquad
\mathbf{g}\in\mathcal{G}_{\mathrm{obs}},
\label{eq:kspace_forward}
\end{equation}
where $\mathbf{M}_{\mathbf{g}}$ is the direction-dependent sampling mask,
$\mathcal{F}_{xy}$ denotes the slice-wise in-plane Fourier transform, and
$\boldsymbol{\epsilon}_{\mathbf{g}}$ represents noise.
Masks may vary across observed directions to provide complementary
spatial-frequency measurements.

Our goal is to recover a continuous field
$\widehat S(\mathbf{x},\mathbf{g})$ from the undersampled measurements of
the observed directions, while maintaining consistency with the acquired
k-space samples.
The estimated field can then be queried at unobserved diffusion directions.
The held-out direction set $\mathcal{G}_{\mathrm{miss}}$ is excluded from
optimization and used only for retrospective evaluation.

\subsubsection{Shared Gaussian Support and Primitive-Centered Aggregation}
\label{sec:primitive_aggregation}

All diffusion directions are represented using a common collection of
$N$ Gaussian primitives.
Primitive $i$ is parameterized by a center
$\boldsymbol{\mu}_i$, a spatial covariance
$\boldsymbol{\Sigma}^{x}_i$, and q-response parameters
$\boldsymbol{\vartheta}_i$.
Its spatial contribution is
\begin{equation}
G_i(\mathbf{x})
=
\exp
\left[
-\frac{1}{2}
(\mathbf{x}-\boldsymbol{\mu}_i)^{\top}
(\boldsymbol{\Sigma}^{x}_i)^{-1}
(\mathbf{x}-\boldsymbol{\mu}_i)
\right].
\label{eq:spatial_gaussian}
\end{equation}
We use a diagonal spatial covariance in the current implementation, allowing
the support of each primitive to adapt independently along the three spatial
axes.
Let
$\rho_i(\mathbf{g},\boldsymbol{\vartheta}_i)$ denote the response amplitude
of primitive $i$ under direction $\mathbf{g}$,
the spatial--angular field is rendered as
\begin{equation}
\widehat S(\mathbf{x},\mathbf{g})
=
\sum_{i=1}^{N}
G_i(\mathbf{x})
\rho_i(\mathbf{g},\boldsymbol{\vartheta}_i).
\label{eq:spatial_angular_rendering}
\end{equation}
Eq.~\eqref{eq:spatial_angular_rendering} defines a primitive-centered
representation.
At each spatial location, several overlapping primitives jointly determine
the signal.
Conversely, each primitive contributes to multiple neighboring locations and
shares one response function across diffusion directions.
The voxel-level response therefore emerges from a spatial mixture of local
primitive responses rather than from an independent directional parameter
vector attached to each voxel.

\subsubsection{Tensor-Anchored Continuous Diffusion Response}
\label{sec:tensor_anchored_response}

To prevent an unconstrained angular model from absorbing spatial
reconstruction errors, each primitive is assigned a compact response anchored
by diffusion-tensor attenuation.
The tensor response is
\begin{equation}
\rho_i^{\mathrm{ten}}(\mathbf{g})
=
a_i
\exp
\left(
-b\,\mathbf{g}^{\top}\mathbf{D}_i\mathbf{g}
\right),
\qquad
\mathbf{D}_i
=
\kappa_D\mathbf{L}_i\mathbf{L}_i^{\top},
\label{eq:tensor_anchor}
\end{equation}
where $a_i>0$ is a primitive-level reference amplitude,
$\kappa_D>0$ sets the diffusivity scale, and the factorization of
$\mathbf{D}_i$ guarantees nonnegative apparent diffusivity.

A single tensor may not fully describe partial-volume effects or
departures from ideal tensor behavior.
We introduce a regularized second-order even angular correction
$\mathbf{r}_i^{\top}\boldsymbol{\phi}(\mathbf{g})$, where
$\boldsymbol{\phi}$ is a normalized real even spherical-harmonic basis up to
order two and $\alpha$ controls the correction strength.
The response is defined in the softplus latent domain as
\begin{equation}
\rho_i(\mathbf{g},\boldsymbol{\vartheta}_i)
=
\operatorname{softplus}
\left[
\operatorname{softplus}^{-1}
\left(
\rho_i^{\mathrm{ten}}(\mathbf{g})
\right)
+
\alpha\,
\mathbf{r}_i^{\top}
\boldsymbol{\phi}(\mathbf{g})
\right].
\label{eq:tensor_anchored_response}
\end{equation}
The tensor provides the dominant attenuation geometry, whereas the
even residual models a departure from that anchor.
The softplus mapping guarantees a positive response, and the use of a
quadratic tensor term together with an even angular basis preserves antipodal
symmetry.

The primitive-level tensor does not restrict the signal at a
voxel to a single-tensor model.
Because Eq.~\eqref{eq:spatial_angular_rendering} aggregates
overlapping primitives, the voxel-level q-space signal is a spatially varying
mixture of multiple tensor-anchored responses.
The primitive tensors should be interpreted as local attenuation
anchors rather than as the voxel-wise tensors used to compute downstream FA
and MD maps.

\subsection{Progressive Field Construction and Optimization}
\label{sec:progressive_optimization}

Directly optimizing Gaussian geometry and continuous q-responses from sparse
joint k--q measurements is poorly conditioned.
A reconstruction mismatch may originate from insufficient spatial support,
measurement inconsistency, or genuine directional variation.
We therefore increase the representation complexity progressively from a
q-independent scaffold, to discrete observed-direction responses and finally
to a continuous tensor-anchored field, as illustrated in
Fig.~\ref{fig:method_overview}.

\subsubsection{Anatomy-Aware Scaffold Initialization}
\label{sec:scaffold_initialization}

For each observed direction, the zero-filled reconstruction is obtained by
inverse Fourier transformation with unmeasured k-space coefficients set to
zero.
Their direction-averaged volume is
\begin{equation}
\overline S^{\mathrm{ZF}}(\mathbf{x})
=
\frac{1}{|\mathcal{G}_{\mathrm{obs}}|}
\sum_{\mathbf{g}\in\mathcal{G}_{\mathrm{obs}}}
S_{\mathbf{g}}^{\mathrm{ZF}}(\mathbf{x}).
\label{eq:zf_observed_mean}
\end{equation}
We first fit a q-independent Gaussian field:
\begin{equation}
\widehat S_{\mathrm{base}}(\mathbf{x})
=
\sum_{i=1}^{N_0}
\overline\rho_i G_i(\mathbf{x}),
\label{eq:base_gaussian_field}
\end{equation}
where $\overline\rho_i$ is a direction-independent amplitude.
This stage initializes Gaussian centers, spatial extents, and
amplitudes before tensor or angular residuals are introduced.
The averaged zero-filled image serves only as an observed-data-derived
structural reference, not as a fully sampled supervision target.

\subsubsection{Observed-Response Calibration and Adaptive Allocation}
\label{sec:adaptive_primitive_allocation}

The initial scaffold may provide insufficient support near tissue boundaries,
fine structures, or regions with complex observed directional variation.
To expose these regions without prematurely imposing a continuous q-response,
we temporarily assign an independent positive amplitude
$\rho^{\mathrm{obs}}_{i,\mathbf{g}}$ to primitive $i$ at each observed
direction:
\begin{equation}
\widehat S_{\mathrm{obs}}(\mathbf{x},\mathbf{g})
=
\sum_{i=1}^{N}
\rho^{\mathrm{obs}}_{i,\mathbf{g}}
G_i(\mathbf{x}),
\qquad
\mathbf{g}\in\mathcal{G}_{\mathrm{obs}}.
\label{eq:observed_response_rendering}
\end{equation}
These amplitudes are nonparametric samples of the local responses supported
by the current scaffold and do not define signals at held-out directions.

We summarize the local need for additional primitive capacity as
\begin{equation}
\eta(\mathbf{x})
=
\left[
\omega_{\mathrm{spa}}\eta_{\mathrm{spa}}(\mathbf{x})
+
\omega_{\mathrm{meas}}\eta_{\mathrm{meas}}(\mathbf{x})
+
\omega_{\mathrm{ang}}\eta_{\mathrm{ang}}(\mathbf{x})
\right]
\mathbf{m}(\mathbf{x}),
\label{eq:allocation_score}
\end{equation}
where $\mathbf{m}$ is the brain mask.
The three terms quantify, respectively, the discrepancy between rendered
fields and observed zero-filled images, sampled k-space residuals
back-projected to image space, and angular variation across observed signals
and primitive amplitudes.
All terms are computed exclusively from
$\mathcal{G}_{\mathrm{obs}}$.

\begin{figure}[!t]
\centering
\includegraphics[width=\columnwidth]
{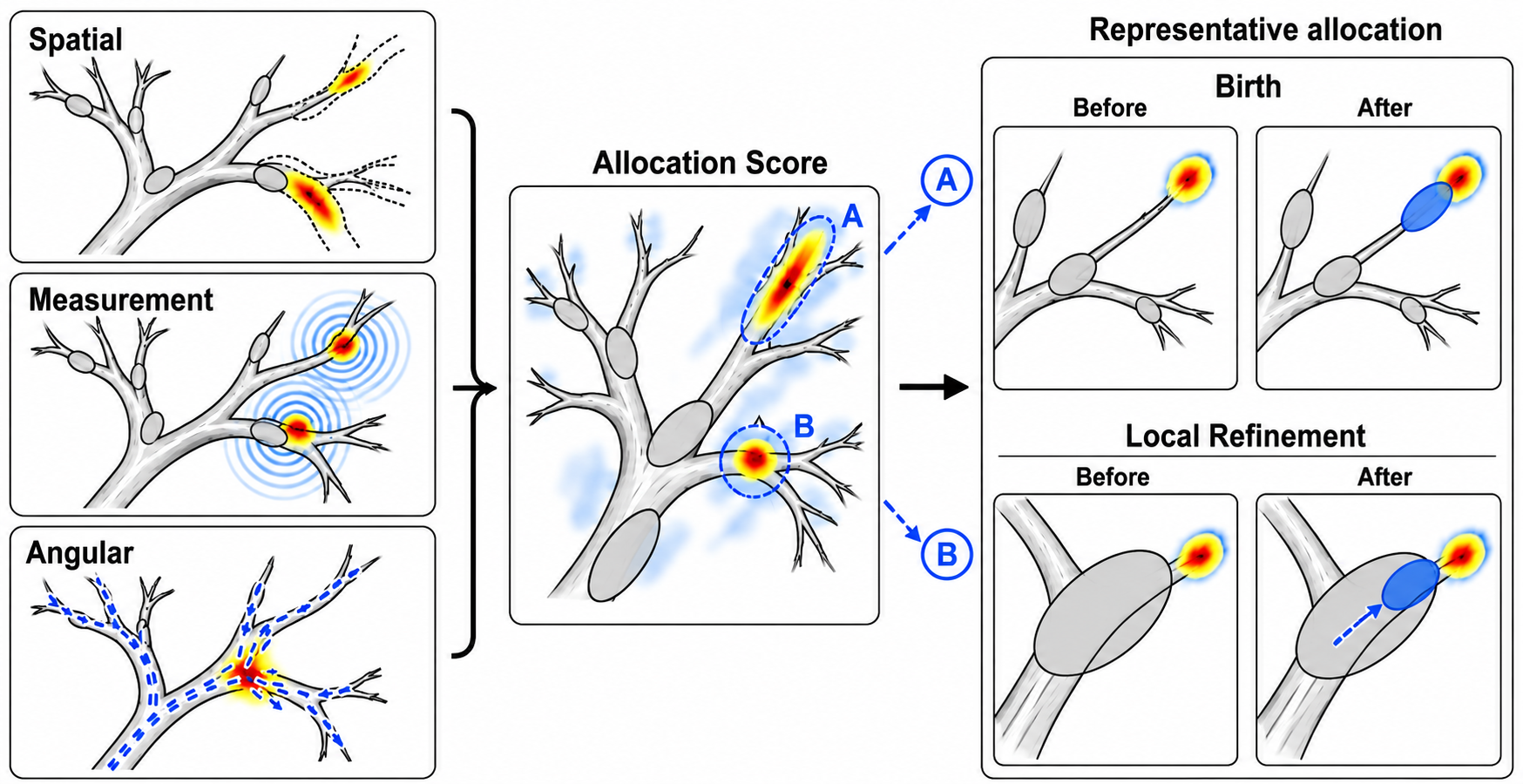}
\caption{
Observed-driven adaptive primitive allocation.
Spatial discrepancy, back-projected k-space residuals, and angular
heterogeneity form an allocation score that guides voxel birth and
primitive splitting, followed by observed-direction response recalibration.
}
\label{fig:adaptive_allocation}
\end{figure}

Primitive capacity is increased through two complementary operations.
Voxel birth introduces new primitives at high-score locations with
insufficient support, whereas primitive splitting locally refines existing
high-score primitives using smaller spatial extents.
The observed-direction amplitudes are recalibrated after allocation, and the
allocation--calibration process is repeated for several rounds.
This strategy concentrates representation capacity in difficult regions
without uniformly increasing the primitive number throughout the volume.
As illustrated in Fig.~\ref{fig:adaptive_allocation}, the allocation process
iteratively identifies under-represented regions and refines the Gaussian
support through birth, splitting, and recalibration.

\subsubsection{Observed-to-Continuous Response Projection and Joint Refinement}
\label{sec:response_projection}

After adaptive allocation, the calibrated amplitudes provide discrete
directional samples for each primitive:
\begin{equation}
\mathcal R_i^{\mathrm{obs}}
=
\left\{
(\mathbf{g},\rho^{\mathrm{obs}}_{i,\mathbf{g}})
\mid
\mathbf{g}\in\mathcal{G}_{\mathrm{obs}}
\right\}.
\label{eq:primitive_response_samples}
\end{equation}
We project these samples onto the continuous response family in
Eq.~\eqref{eq:tensor_anchored_response}:
\begin{equation}
\boldsymbol{\vartheta}_i^{(0)}
=
\argmin_{\boldsymbol{\vartheta}_i}
\sum_{\mathbf{g}\in\mathcal{G}_{\mathrm{obs}}}
\left|
\rho_i(\mathbf{g},\boldsymbol{\vartheta}_i)
-
\rho^{\mathrm{obs}}_{i,\mathbf{g}}
\right|^2
+
\beta_r\|\mathbf r_i\|_2^2.
\label{eq:response_projection_objective}
\end{equation}
The reference amplitude is initialized from a robust upper
statistic of the observed primitive amplitudes, the tensor is initialized
from an isotropic diffusivity estimate, and the angular residual is obtained
using a regularized low-order fit.
These quantities are used only for initialization and are
optimized through the nonlinear response in
Eq.~\eqref{eq:tensor_anchored_response}.

The projection converts flexible but discrete observed-direction amplitudes
into a compact, positive, antipodally symmetric function that can be evaluated
at arbitrary directions.
The response parameters are then jointly refined with conservative updates to
the Gaussian centers and spatial extents.
A weak primitive-response anchor maintains consistency with the calibrated
observed amplitudes while allowing interpolation between observed directions.

\subsubsection{Self-Supervised Learning Objectives}
\label{sec:self_supervised_objectives}

The progressive levels share an observed-data-only supervision principle but
use objectives matched to their representation complexity.
The rendered signal is restricted to the brain support before Fourier
encoding.

For an observed direction, sampled k-space consistency is defined as
\begin{equation}
\mathcal L_k
=
\frac{1}{|\mathcal G_{\mathrm{obs}}|}
\sum_{\mathbf g\in\mathcal G_{\mathrm{obs}}}
\frac{
\left\|
\mathbf M_{\mathbf g}
\odot
\left[
\mathcal F_{xy}
\left(
\mathbf m\odot\widehat S(\cdot,\mathbf g)
\right)
-
\mathbf y_{\mathbf g}
\right]
\right\|_1
}{
\|\mathbf M_{\mathbf g}\|_1
}.
\label{eq:kspace_data_consistency}
\end{equation}
The scaffold initialization minimizes a masked image-domain discrepancy to
$\overline S^{\mathrm{ZF}}$ with weak spatial regularization.
The observed-response calibration uses sampled k-space consistency together
with an auxiliary zero-filled anchor.
The final continuous-field objective is
\begin{align}
\mathcal L_{\mathrm{field}}
={}&
\mathcal L_k
+
\lambda_{\mathrm{ZF}}\mathcal L_{\mathrm{ZF}}
+
\lambda_{\mathrm{TV}}\mathcal L_{\mathrm{TV}}
+
\lambda_{\rho}\mathcal L_{\rho}
\nonumber\\
&+
\lambda_D\mathcal L_D
+
\lambda_r\mathcal L_r ,
\label{eq:final_field_objective}
\end{align}
where $\mathcal L_{\mathrm{ZF}}$ weakly anchors the rendered observed
directions to their zero-filled reconstructions,
$\mathcal L_{\mathrm{TV}}$ regularizes local image variation, and
\begin{equation}
\mathcal L_{\rho}
=
\frac{1}{N|\mathcal G_{\mathrm{obs}}|}
\sum_{i=1}^{N}
\sum_{\mathbf g\in\mathcal G_{\mathrm{obs}}}
\left|
\rho_i(\mathbf g,\boldsymbol{\vartheta}_i)
-
\rho^{\mathrm{obs}}_{i,\mathbf g}
\right|
\label{eq:response_anchor_loss}
\end{equation}
transfers the locally calibrated observed responses to the continuous field.
The terms $\mathcal L_D$ and $\mathcal L_r$ regularize the tensor parameters
and angular correction, respectively.
Algorithm~\ref{alg:progressive_gaussian_field} summarizes the complete
subject-specific reconstruction procedure.
The optimization proceeds from a q-independent spatial scaffold, through
observed-direction response calibration and adaptive primitive allocation,
to continuous-response projection and joint field refinement.
Only sampled k-space measurements from
$\mathcal G_{\mathrm{obs}}$ are used throughout the procedure.

\begin{algorithm}[!t]
\caption{Progressive Gaussian Field Optimization}
\label{alg:progressive_gaussian_field}
\footnotesize

\noindent\textbf{Input:}
Sampled measurements
$\{\mathbf y_{\mathbf g},\mathbf M_{\mathbf g}\}_{
\mathbf g\in\mathcal G_{\mathrm{obs}}}$,
brain mask $\mathbf m$, initial primitive number $N_0$,
and allocation rounds $R_{\mathrm{alloc}}$

\noindent\textbf{Output:}
Spatial--angular field $\widehat S(\mathbf x,\mathbf g)$

\begin{algorithmic}[1]

\ForAll{$\mathbf g\in\mathcal G_{\mathrm{obs}}$}
    \State
    $S_{\mathbf g}^{\mathrm{ZF}}
    \gets
    \mathcal F_{xy}^{-1}(\mathbf y_{\mathbf g})$
\EndFor

\State
$\overline S^{\mathrm{ZF}}
\gets
|\mathcal G_{\mathrm{obs}}|^{-1}
\sum_{\mathbf g\in\mathcal G_{\mathrm{obs}}}
S_{\mathbf g}^{\mathrm{ZF}}$

\State Initialize $N_0$ shared Gaussian primitives from
$\overline S^{\mathrm{ZF}}$

\State Optimize the q-independent Gaussian scaffold

\State Initialize observed-direction responses
$\{\rho^{\mathrm{obs}}_{i,\mathbf g}\}$

\For{$r=1$ to $R_{\mathrm{alloc}}$}
    \State Calibrate
    $\{\rho^{\mathrm{obs}}_{i,\mathbf g}\}$
    using sampled k-space consistency

    \State Compute allocation score $\eta$

    \State Add primitives by voxel birth and local splitting

    \State Recalibrate observed responses on the updated support
\EndFor

\For{$i=1$ to $N$}
    \State
    $\mathcal R_i^{\mathrm{obs}}
    \gets
    \{(\mathbf g,\rho^{\mathrm{obs}}_{i,\mathbf g})
    \mid
    \mathbf g\in\mathcal G_{\mathrm{obs}}\}$

    \State Fit
    $\boldsymbol{\vartheta}_i$
    from $\mathcal R_i^{\mathrm{obs}}$
    using the tensor--residual response
\EndFor

\State Jointly refine Gaussian support and
$\{\boldsymbol{\vartheta}_i\}_{i=1}^{N}$
with $\mathcal L_{\mathrm{field}}$

\State \Return
$\widehat S(\mathbf x,\mathbf g)
=
\sum_{i=1}^{N}
G_i(\mathbf x)
\rho_i(\mathbf g,\boldsymbol{\vartheta}_i)$

\end{algorithmic}
\end{algorithm}

\section{Experiments}
\label{sec:experiments}

\subsection{Experimental Setup}
\label{sec:exp_setup}

\subsubsection{Dataset}

We evaluated the proposed method on retrospectively undersampled diffusion
MRI data from the Human Connectome Project (HCP), which provides
high-angular-resolution acquisitions across multiple diffusion
shells~\cite{r_hcp2013,r_hcp2016}. The shells with $b$-values of 1000, 2000,
and $3000~\mathrm{s/mm^2}$ were evaluated independently. The DWIs were
normalized by the mean $b=0$ image and cropped to the brain mask
bounding box with a four-voxel margin. Reconstruction and
evaluation were performed within the brain mask using the same preprocessing,
crop, and mask for all methods.

For each shell, the fully sampled DWI volumes were transformed slice by slice
into single-coil-equivalent in-plane k-space. Spatial undersampling was
simulated using direction-dependent variable-density one-dimensional
Cartesian masks with a fully sampled central fraction of 0.08. We evaluated
$R_k\in\{3,4\}$ with
$N_{\mathrm{obs}}\in\{30,15,10\}$ observed directions, corresponding to
$R_q=90/N_{\mathrm{obs}}\in\{3,6,9\}$. The remaining directions were held out
for angular reconstruction evaluation, yielding six joint acceleration
configurations per shell. Identical direction splits and k-space masks were
used for all methods under each condition.

\subsubsection{Comparison Methods}

The proposed method was compared with six representative baselines covering
sequential spatial--angular reconstruction, joint k--q model-based
reconstruction, and Gaussian-based decoupled reconstruction.
The sequential spatial--angular pipelines included zero-filled reconstruction
followed by spherical-harmonic interpolation (ZF+SH) and compressed-sensing
reconstruction with total-variation regularization followed by
spherical-harmonic interpolation (CSTV+SH).
ZF+SH directly applied inverse Fourier reconstruction to the observed
directions before angular interpolation, whereas CSTV+SH first reconstructed
each observed DWI using a conventional compressed-sensing MRI formulation
and then synthesized the missing directions in the spherical-harmonic
domain~\cite{r_lustig2007}.
The joint k--q model-based methods included JointKQ-CS, which exploits joint
spatial--angular sparsity~\cite{r_kqcs2017}, and qModeL-DAE, which incorporates
a learned q-space prior into a measurement-consistent reconstruction
framework~\cite{r_qmodel2020}.
The Gaussian-based decoupled baselines included 3DGS+SH and 3DGS+PCCNN.
Both reconstructed the observed-direction DWIs using an explicit
3D Gaussian representation adapted from Gaussian-based MRI
reconstruction~\cite{r_3dgsmr2025}.
Missing directions were subsequently estimated using spherical-harmonic
interpolation or PCCNN-based spatial--angular
super-resolution~\cite{r_pccnn2023}, respectively.

All methods were evaluated using the same crop, brain mask, observed
directions, held-out directions, and k-space undersampling masks.
For subject-specific methods, the held-out directions were
excluded from all optimization objectives and were used only for retrospective
evaluation.

\subsubsection{Evaluation Metrics}

Missing-direction reconstruction was evaluated using volumetric peak
signal-to-noise ratio (PSNR) and structural similarity index measure (SSIM)
within the three-dimensional brain mask, with results averaged over all
held-out directions. Diffusion tensors were fitted from the completed
direction sets using the same procedure for all methods, and the resulting
FA and MD maps were evaluated using PSNR and SSIM. Local orientation accuracy was measured by the angular error
between the reference and reconstructed principal eigenvectors (PEVs):
\begin{equation}
e_{\mathrm{PEV}}(\mathbf{x})
=
\frac{180}{\pi}
\cos^{-1}\!\left(
\left|
\mathbf{v}_{1}^{\mathrm{ref}}(\mathbf{x})^{\top}
\mathbf{v}_{1}^{\mathrm{rec}}(\mathbf{x})
\right|
\right),
\label{eq:pev_error}
\end{equation}
where the absolute inner product removes the sign ambiguity of tensor
eigenvectors. Tensor-glyph visualizations were also used to examine local
anisotropy and orientation consistency.

\subsubsection{Implementation Details}

The proposed method was implemented in PyTorch and optimized separately for
each subject and diffusion shell. The representation was initialized with
approximately $1.4\times10^{5}$ Gaussian primitives and increased to
approximately $2.4\times10^{5}$ through adaptive densification. Each
primitive contained a learnable three-dimensional center, diagonal spatial
covariance, positive-semidefinite diffusion tensor, and regularized
second-order even angular residual. The brain mask was applied during
rendering and evaluation. All experiments were conducted using an NVIDIA
GeForce RTX 4090 GPU and an Intel Xeon CPU. The source code is publicly available at
\url{https://github.com/yqx7150/DEGF}.

\subsection{Comparison}
\label{sec:results}

\subsubsection{DWI Reconstruction and Missing-Direction Synthesis}
\label{sec:res_dwi}

Table~\ref{tab:main_multisetting} reports missing-direction DWI reconstruction
across the three HCP shells and joint k--q acceleration settings. All methods
used identical masks and direction splits, and PSNR/SSIM were computed over
held-out directions within the brain mask. The proposed method consistently achieves the best missing-direction reconstruction performance across all tested $b$-values and acceleration settings.
The advantage becomes more evident under stronger joint acceleration, such as 15 or 10 observed diffusion directions combined with $R_k=4$.
This indicates that the primitive-level tensor-residual response provides
more stable angular generalization than two-stage reconstruction followed by
spherical harmonic interpolation or learning-based angular completion.

\begin{figure*}[!t]
\centering
\includegraphics[width=1.028\textwidth]{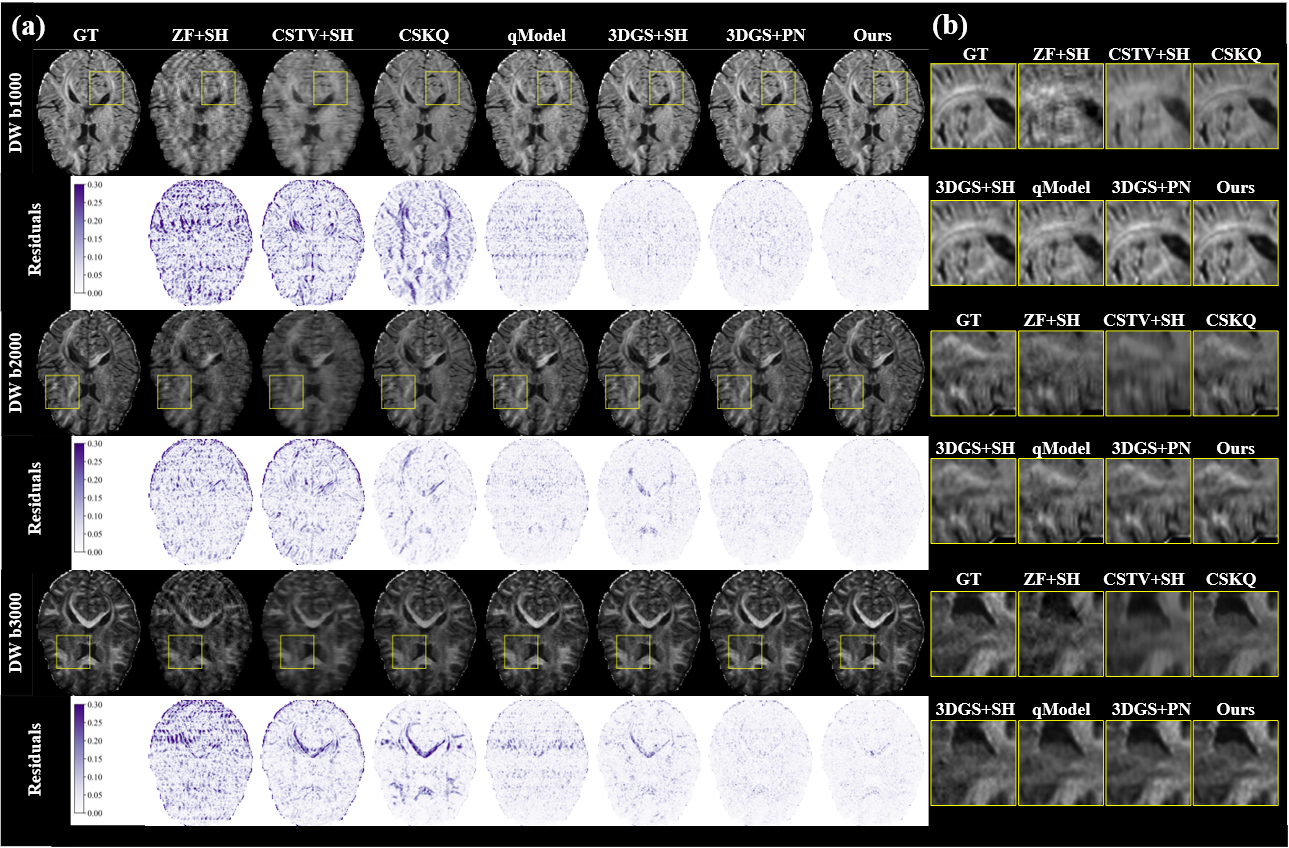}
\caption{
Missing-direction DWI reconstruction at $R_k=3$ with 15 observed directions.
(a) Reconstructions and absolute-error maps for the three diffusion shells, (b) enlarged views of the highlighted regions.
Columns follow the displayed method order.
}
\label{fig:main_multib_visual}
\end{figure*}

\begin{table*}[!t]
\centering
\caption{Missing-direction DWI reconstruction on HCP dataset under
different spatial and angular acceleration settings.}
\label{tab:main_multisetting}

\begin{threeparttable}
\tiny
\setlength{\tabcolsep}{1.55pt}
\renewcommand{\arraystretch}{0.88}

\resizebox{\textwidth}{!}{%
\begin{tabular}{ccccccccccc}
\toprule
$b$-value & $R_k$ & Obs. & $R_q$
& ZF+SH
& CSTV+SH
& JointKQ-CS
& qModeL-DAE
& 3DGS+SH
& 3DGS+PCCNN
& Ours \tabularnewline
\midrule

$1000$ & 3 & 30 & 3
& 22.92/0.8754
& 22.36/0.5944
& 30.18/0.9728
& 30.51/0.9701
& \underline{30.94}/0.9752
& 30.76/\underline{0.9760}
& \textbf{33.59}/\textbf{0.9851}
\tabularnewline

$1000$ & 3 & 15 & 6
& 19.30/0.7495
& 21.99/0.5723
& 27.28/0.9583
& 27.79/0.9441
& \underline{30.24}/0.9694
& 29.94/\underline{0.9695}
& \textbf{32.58}/\textbf{0.9816}
\tabularnewline

$1000$ & 3 & 10 & 9
& 21.43/0.8491
& 21.52/0.5571
& 24.90/0.9401
& 28.73/0.9545
& \underline{29.57}/\underline{0.9638}
& 28.95/0.9617
& \textbf{30.91}/\textbf{0.9733}
\tabularnewline

$1000$ & 4 & 30 & 3
& 21.47/0.8287
& 20.98/0.5225
& \underline{28.99}/\underline{0.9610}
& 27.51/0.9410
& 28.79/0.9572
& 28.77/0.9589
& \textbf{33.16}/\textbf{0.9834}
\tabularnewline

$1000$ & 4 & 15 & 6
& 18.67/0.7191
& 20.83/0.5097
& 26.38/0.9414
& 25.39/0.9079
& \underline{27.89}/0.9460
& 27.86/\underline{0.9472}
& \textbf{31.79}/\textbf{0.9773}
\tabularnewline

$1000$ & 4 & 10 & 9
& 20.24/0.7994
& 20.53/0.4972
& 24.22/0.9181
& 25.93/0.9171
& \underline{26.73}/0.9303
& 26.67/\underline{0.9308}
& \textbf{29.97}/\textbf{0.9651}
\tabularnewline

\midrule

$2000$ & 3 & 30 & 3
& 24.68/0.8809
& 23.66/0.5936
& 31.01/0.9691
& 31.15/0.9658
& 30.75/0.9665
& \underline{31.41}/\underline{0.9704}
& \textbf{34.18}/\textbf{0.9813}
\tabularnewline

$2000$ & 3 & 15 & 6
& 18.78/0.6693
& 22.81/0.5682
& 28.69/0.9535
& 28.79/0.9429
& 30.45/0.9626
& \underline{30.98}/\underline{0.9651}
& \textbf{33.01}/\textbf{0.9762}
\tabularnewline

$2000$ & 3 & 10 & 9
& 22.15/0.8270
& 22.28/0.5431
& 25.91/0.9278
& 29.07/0.9476
& 29.83/0.9573
& \underline{29.86}/\underline{0.9588}
& \textbf{31.00}/\textbf{0.9644}
\tabularnewline

$2000$ & 4 & 30 & 3
& 23.16/0.8315
& 22.40/0.5240
& \underline{29.90}/\underline{0.9580}
& 28.56/0.9398
& 29.14/0.9503
& 29.44/0.9516
& \textbf{33.66}/\textbf{0.9790}
\tabularnewline

$2000$ & 4 & 15 & 6
& 18.72/0.6572
& 21.76/0.5055
& 27.73/0.9367
& 26.58/0.9085
& 28.79/0.9436
& \underline{29.05}/\underline{0.9437}
& \textbf{32.23}/\textbf{0.9712}
\tabularnewline

$2000$ & 4 & 10 & 9
& 21.29/0.7874
& 21.61/0.4950
& 25.25/0.9063
& 26.77/0.9128
& 27.74/0.9295
& \underline{27.85}/\underline{0.9314}
& \textbf{30.42}/\textbf{0.9581}
\tabularnewline

\midrule

$3000$ & 3 & 30 & 3
& 25.47/0.8821
& 23.54/0.5887
& \underline{32.26}/\underline{0.9670}
& 31.75/0.9630
& 30.68/0.9568
& 32.01/0.9665
& \textbf{34.36}/\textbf{0.9761}
\tabularnewline

$3000$ & 3 & 15 & 6
& 20.33/0.6786
& 23.05/0.5696
& 30.10/0.9528
& 28.15/0.9199
& 30.49/0.9531
& \underline{31.50}/\underline{0.9607}
& \textbf{33.12}/\textbf{0.9694}
\tabularnewline

$3000$ & 3 & 10 & 9
& 24.32/0.8472
& 22.72/0.5517
& 27.21/0.9290
& 28.97/0.9332
& 29.67/0.9449
& \underline{30.32}/\underline{0.9518}
& \textbf{31.07}/\textbf{0.9546}
\tabularnewline

$3000$ & 4 & 30 & 3
& 23.99/0.8388
& 22.57/0.5257
& \underline{31.27}/\underline{0.9580}
& 29.48/0.9403
& 29.52/0.9424
& 30.24/0.9475
& \textbf{33.89}/\textbf{0.9736}
\tabularnewline

$3000$ & 4 & 15 & 6
& 18.92/0.6291
& 22.24/0.5141
& 29.21/0.9395
& 26.23/0.8803
& 29.23/0.9359
& \underline{29.82}/\underline{0.9404}
& \textbf{32.47}/\textbf{0.9640}
\tabularnewline

$3000$ & 4 & 10 & 9
& 23.14/0.8046
& 22.08/0.5040
& 26.54/0.9097
& 27.06/0.8993
& 28.15/0.9199
& \underline{28.54}/\underline{0.9243}
& \textbf{30.65}/\textbf{0.9484}
\tabularnewline

\bottomrule
\end{tabular}%
}

\end{threeparttable}
\end{table*}

Fig.~\ref{fig:main_multib_visual} presents representative DWI visual comparisons under different diffusion weightings.
The zero-filled and compressed-sensing baselines suffer from residual aliasing artifacts and unstable angular interpolation, especially under high acceleration.
The two-stage Gaussian baselines improve spatial sharpness but do not fully exploit the shared diffusion response across directions.
In contrast, the proposed method better preserves anatomical edges, suppresses undersampling artifacts, and maintains direction-dependent diffusion contrast across different $b$-values.

\subsubsection{Downstream Diffusion Metric and Orientation Evaluation}
\label{sec:res_downstream}

To evaluate whether the reconstructed DWI signals preserve downstream
diffusion information, we further computed downstream diffusion tensor metrics from the reconstructed full-direction datasets.
For each method, diffusion tensors were fitted using the same estimation
protocol, and FA, MD, PEVs, and tensor glyphs were compared with the reference.

\begin{figure*}[!t]
\centering
\includegraphics[width=\textwidth]{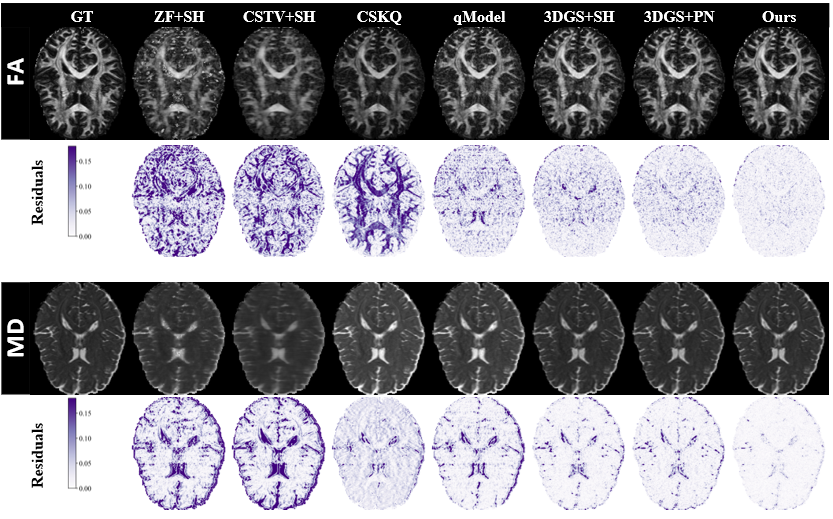}
\caption{
FA and MD maps with absolute errors at $R_k=3$ and 15 observed directions.
Rows show FA, FA error, MD, and MD error. Columns follow the displayed
method order.
}
\label{fig:fa_visual}
\end{figure*}

\begin{figure*}[!t]
\centering

\begin{minipage}[t]{0.242\textwidth}
    \centering
    \includegraphics[width=\linewidth]{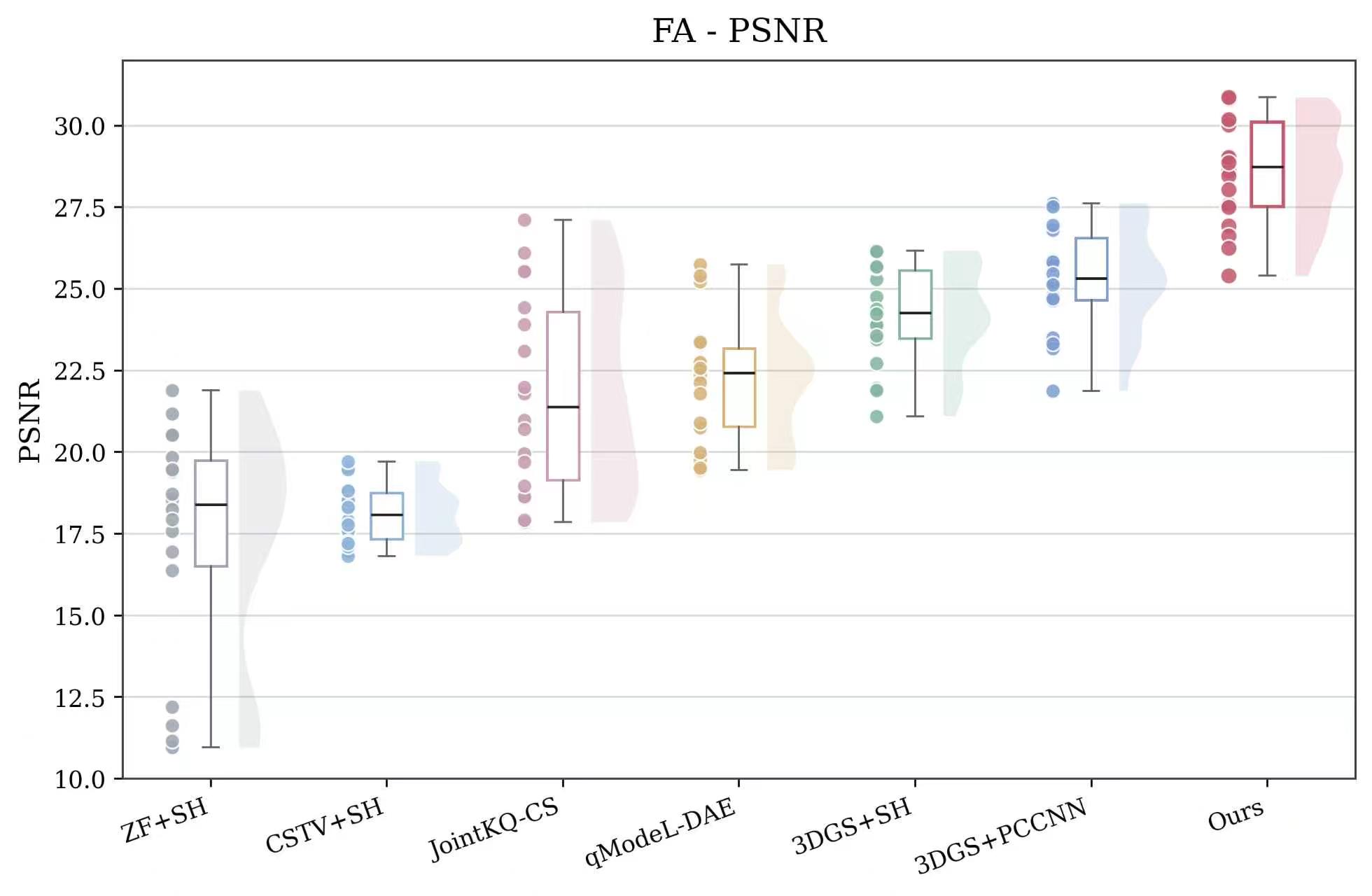}
    \vspace{-1mm}
    \centerline{\scriptsize (a)}
\end{minipage}
\hfill
\begin{minipage}[t]{0.242\textwidth}
    \centering
    \includegraphics[width=\linewidth]{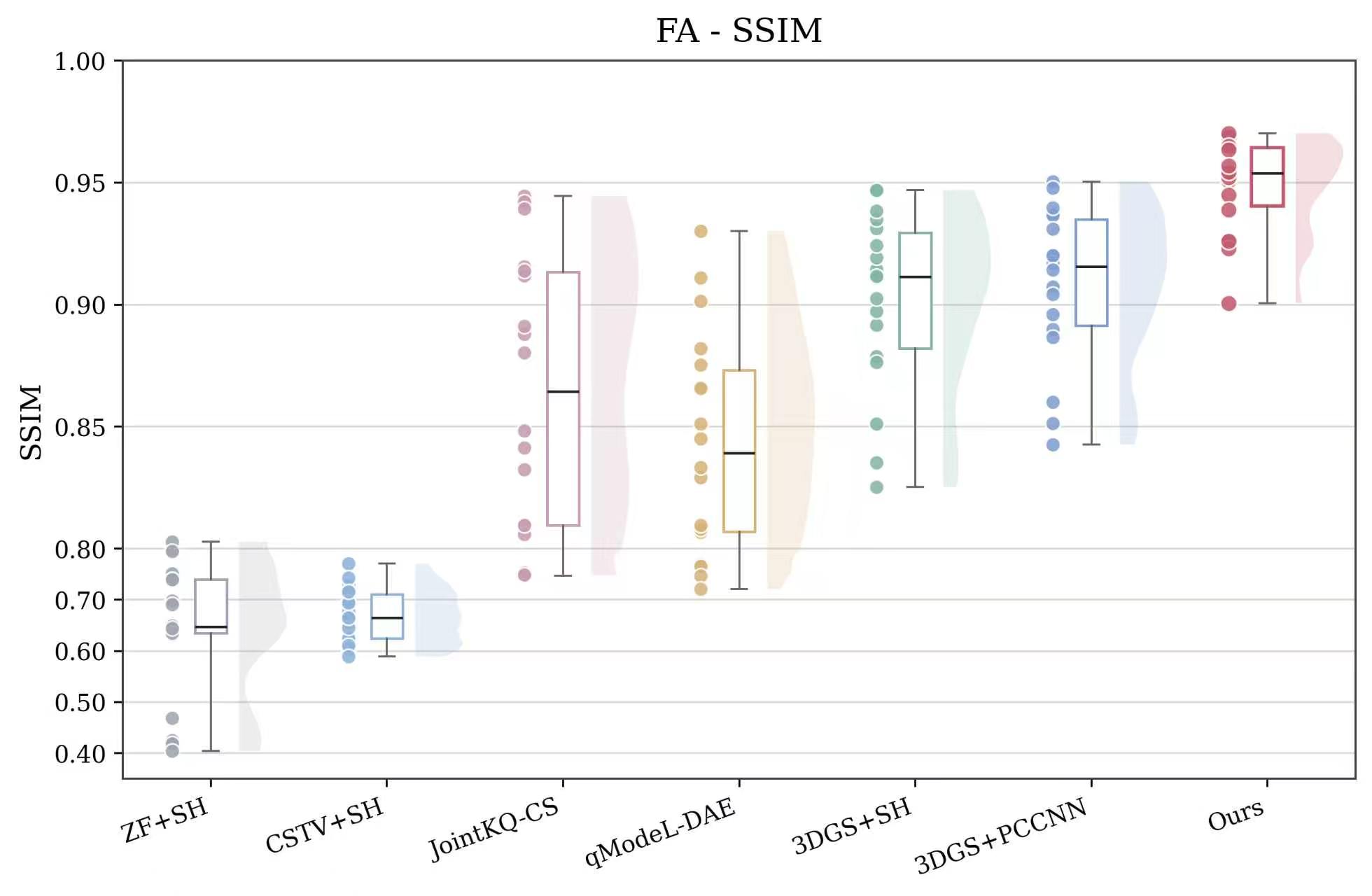}
    \vspace{-1mm}
    \centerline{\scriptsize (b)}
\end{minipage}
\hfill
\begin{minipage}[t]{0.242\textwidth}
    \centering
    \includegraphics[width=\linewidth]{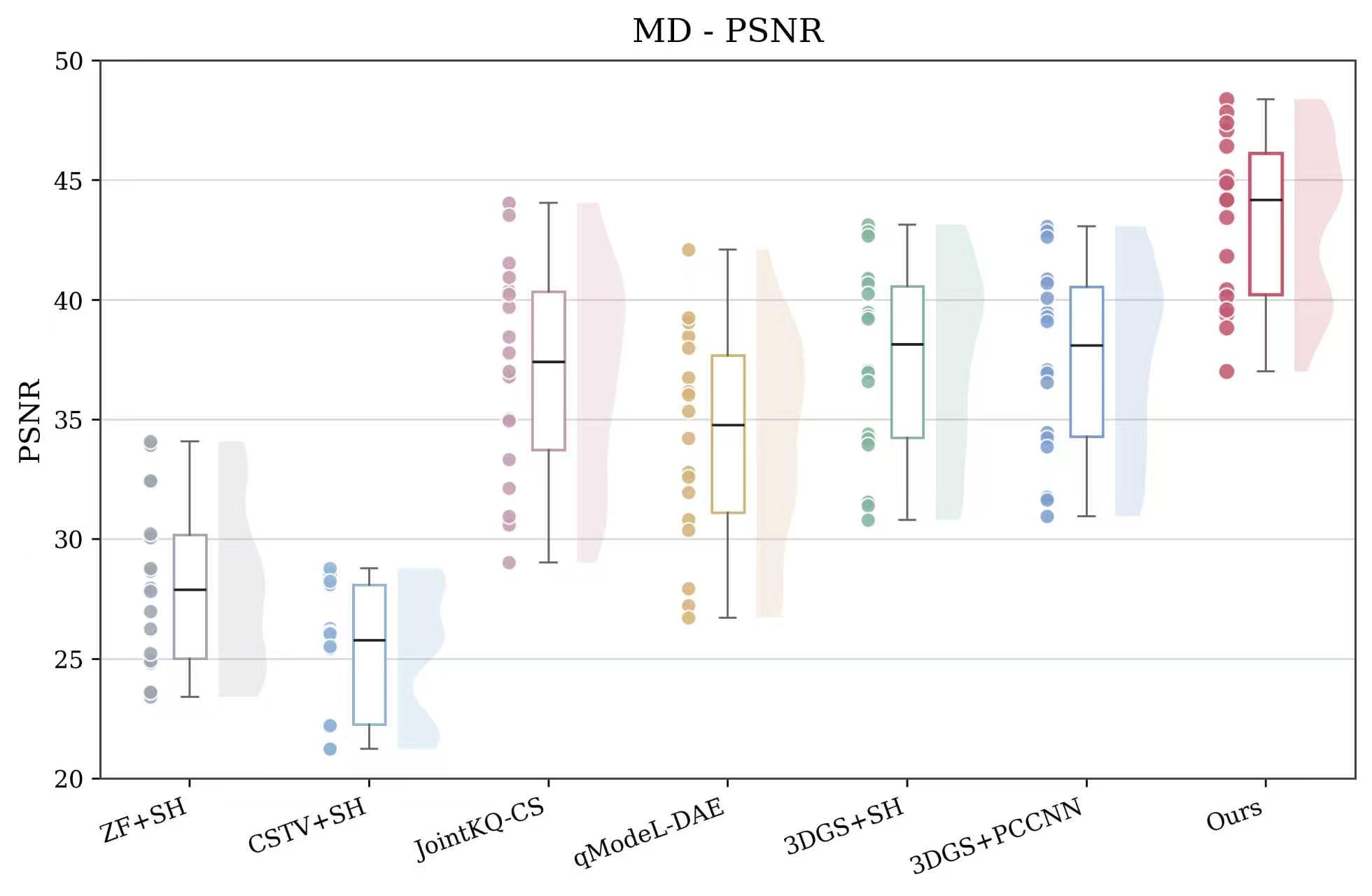}
    \vspace{-1mm}
    \centerline{\scriptsize (c)}
\end{minipage}
\hfill
\begin{minipage}[t]{0.242\textwidth}
    \centering
    \includegraphics[width=\linewidth]{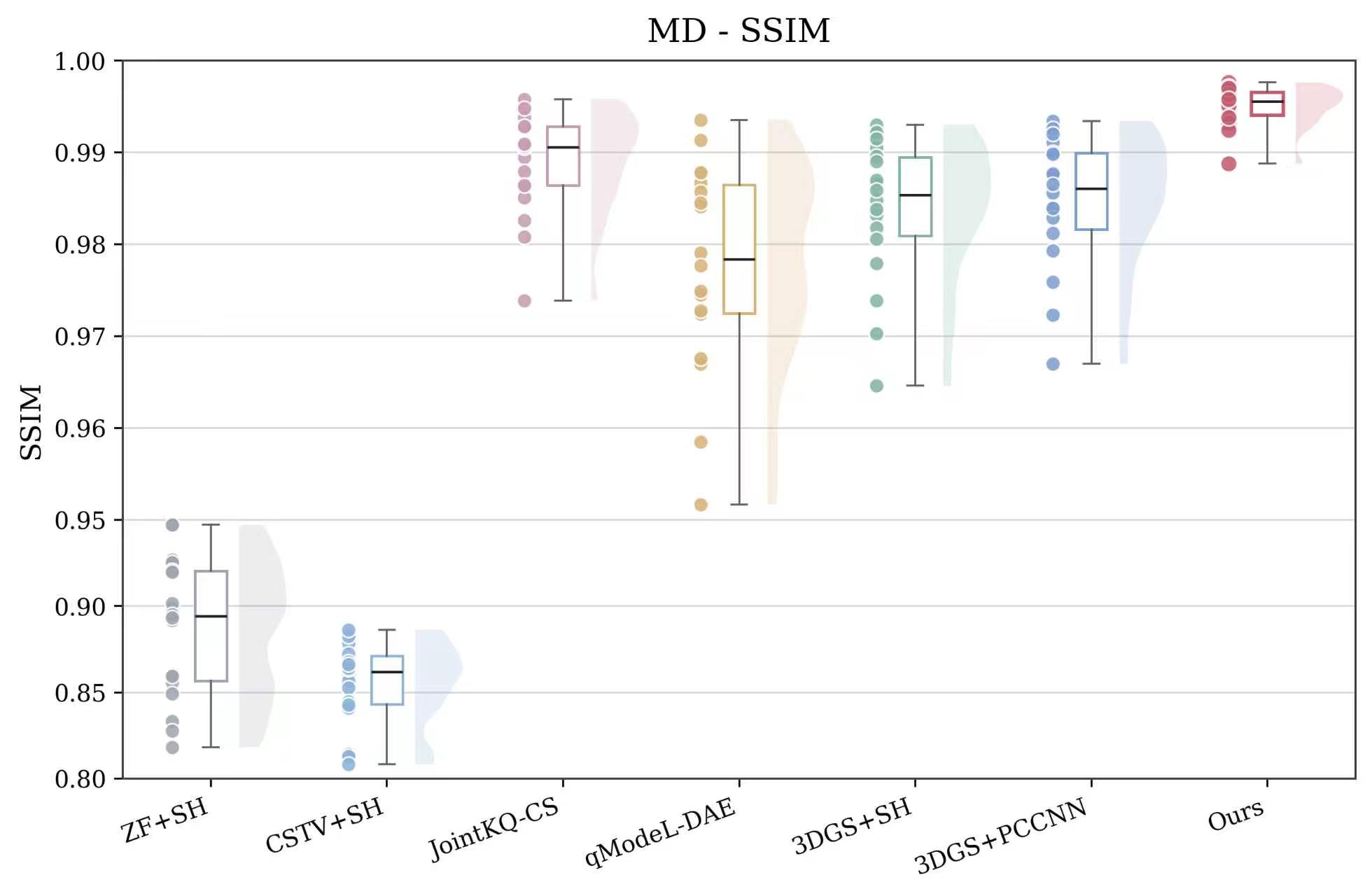}
    \vspace{-1mm}
    \centerline{\scriptsize (d)}
\end{minipage}

\vspace{1mm}
\caption{
FA and MD reconstruction accuracy across all HCP joint-acceleration settings.
(a) FA PSNR, (b) FA SSIM, (c) MD PSNR, and (d) MD SSIM.
}
\label{fig:downstream_metric_summary}
\end{figure*}

Fig.~\ref{fig:fa_visual} shows representative FA and MD maps together with
their absolute-error maps, while Fig.~\ref{fig:downstream_metric_summary}
summarizes the corresponding quantitative results across all evaluated
settings. The proposed method produces sharper FA structures and more faithful
MD contrast than competing methods, particularly in white-matter regions where
angular reconstruction errors can propagate into tensor fitting and degrade
diffusion-metric estimation. This visual improvement is consistent with the
quantitative results, where the proposed method achieves higher FA and MD
reconstruction accuracy across the tested acceleration conditions. These
findings indicate that improved missing-direction DWI reconstruction is
preserved in downstream tensor-derived metrics rather than being limited to
image-level fidelity.

\begin{figure*}[!t]
\centering
\setlength{\tabcolsep}{3pt}
\begin{tabular}{cc}
\includegraphics[width=0.49\textwidth]{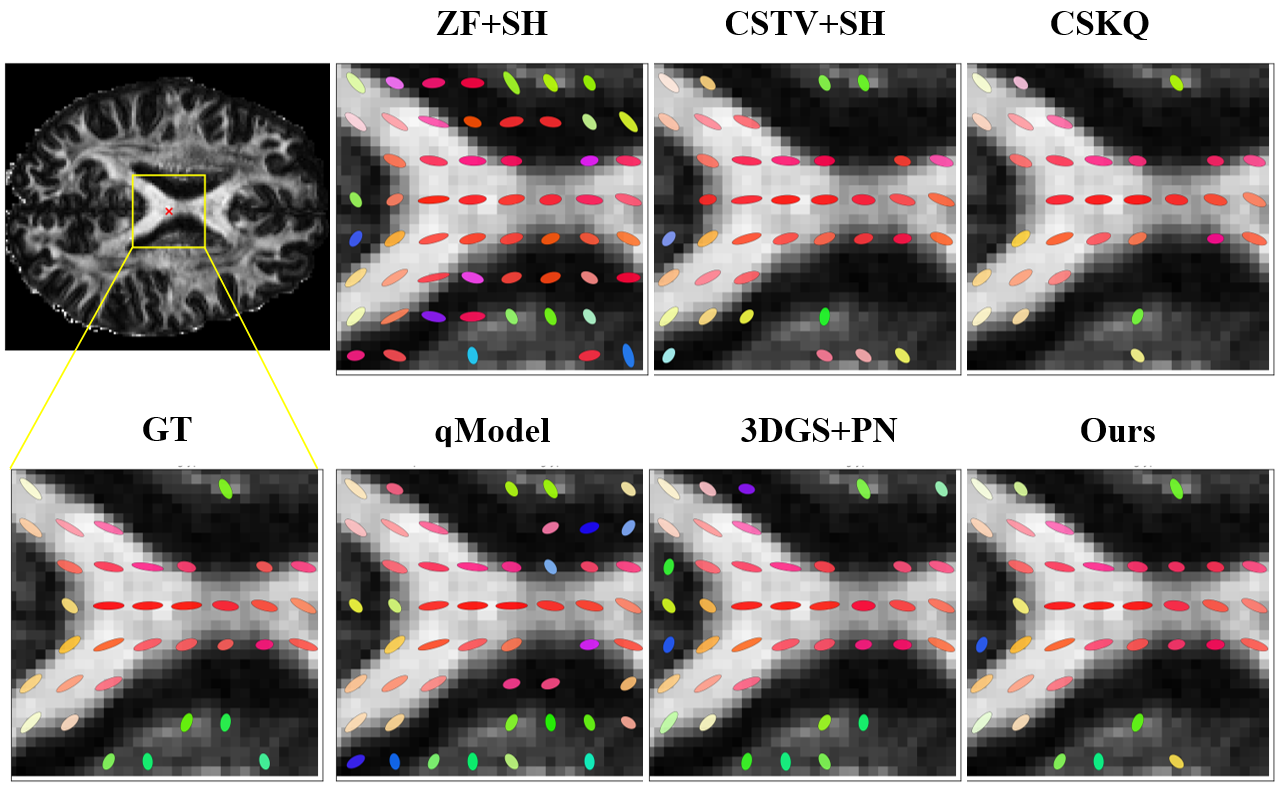} &
\includegraphics[width=0.52\textwidth]{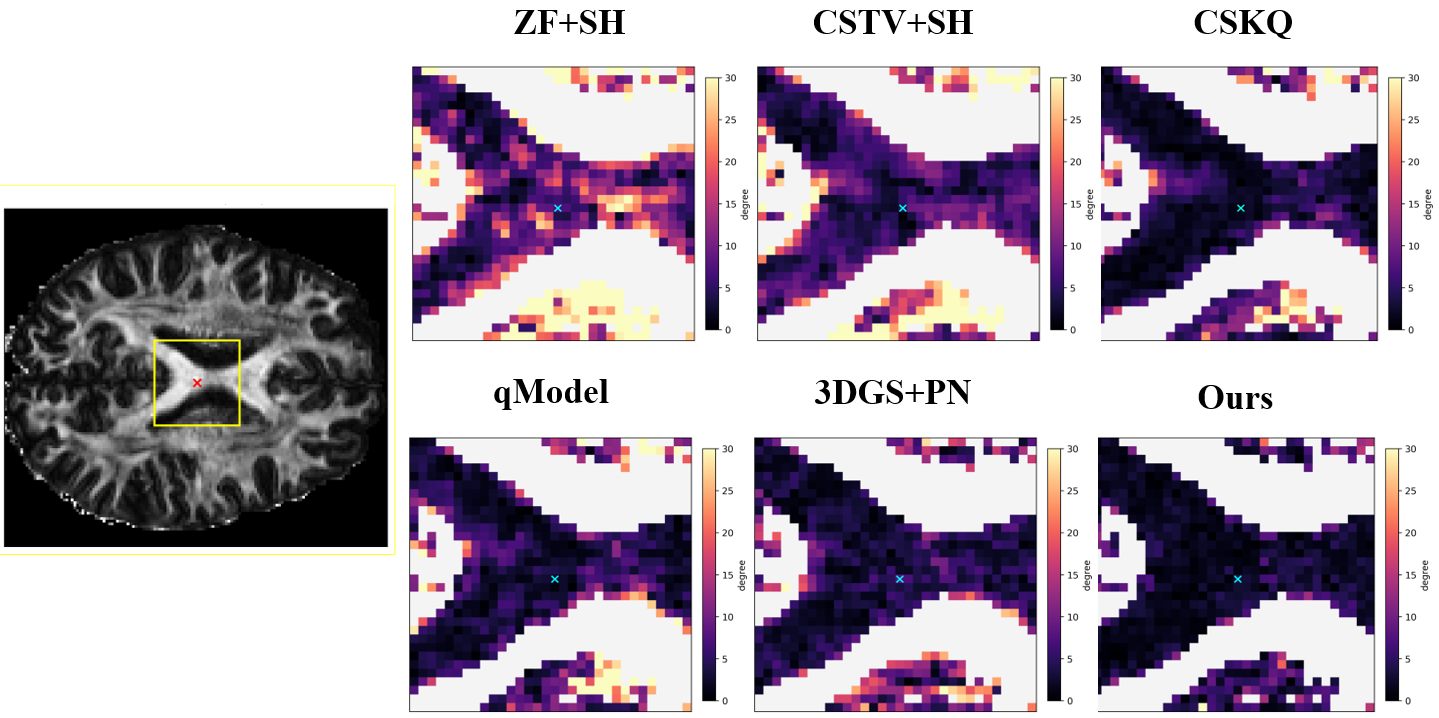} \\
{\small (a)} &
{\small (b)}
\end{tabular}
\caption{
Orientation comparison in a white-matter ROI.
(a) Tensor glyphs on FA maps, (b) PEV error maps.
Warmer colors indicate larger errors.
}
\label{fig:orientation_comparison}
\end{figure*}

Local orientation consistency was further assessed using the PEV angular
error in representative white-matter regions according to
Eq.~\eqref{eq:pev_error}. As shown in
Fig.~\ref{fig:orientation_comparison}(a), the tensor glyphs reconstructed by
the proposed method better preserve the reference orientation patterns.
The PEV angular-error maps in Fig.~\ref{fig:orientation_comparison}(b)
further confirm this observation. Baseline methods exhibit larger orientation
deviations in white-matter regions, whereas the proposed method maintains
lower angular error and more continuous local directional structure. These
results demonstrate that the proposed primitive-level diffusion response
improves both scalar diffusion metrics and orientation-sensitive tensor
measurements.

\subsubsection{Analysis of Gaussian Optimization Steps and Primitive Number}
\label{sec:res_gaussian_analysis}

\begin{figure}[!t]
\centering
\includegraphics[width=\columnwidth]{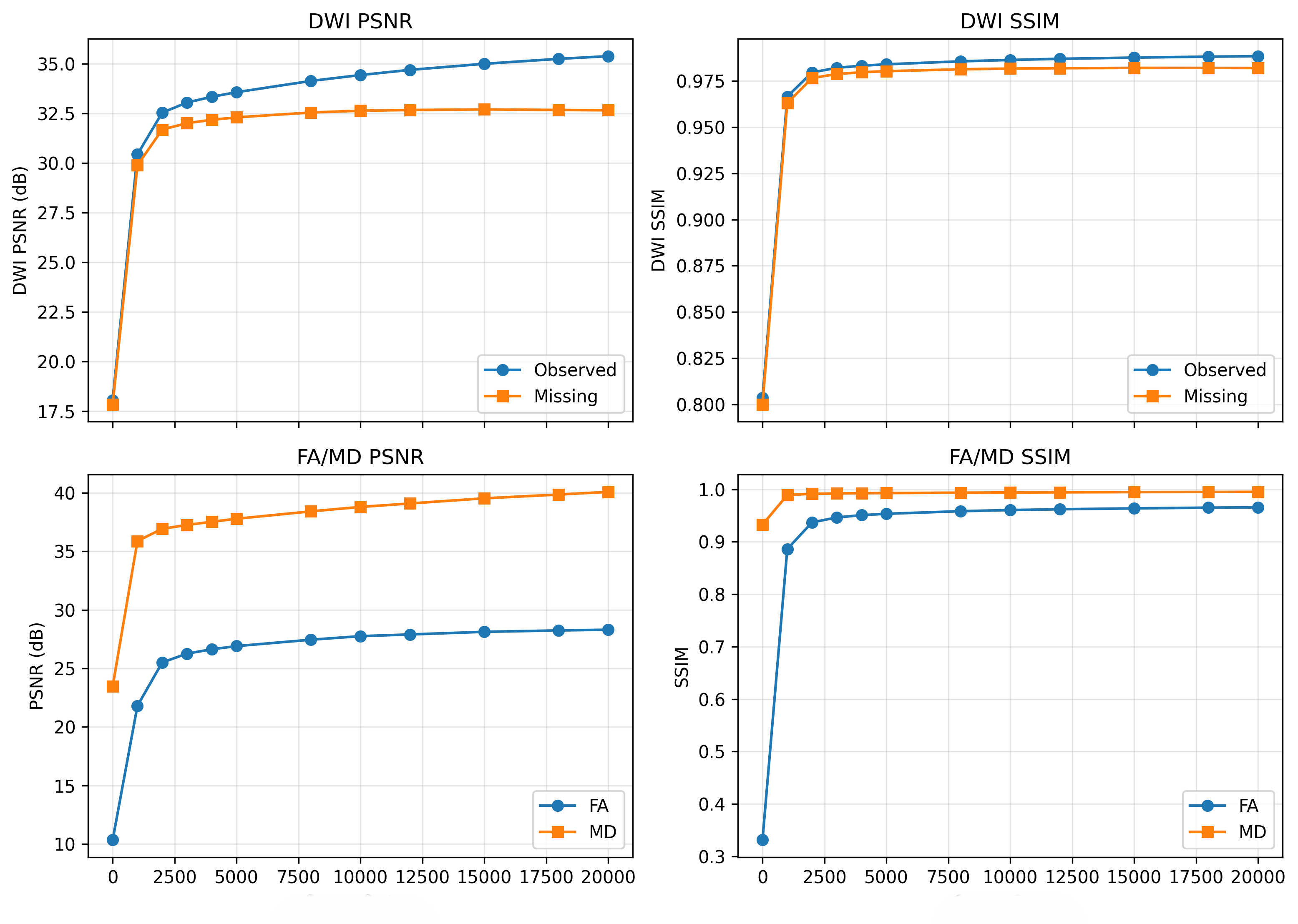}
\caption{
Effect of optimization steps at $b=1000~\mathrm{s/mm^2}$,
$R_k=3$, with 15 observed directions.
}
\label{fig:gaussian_steps}
\end{figure}

We further investigated the influence of Gaussian optimization steps and the number of Gaussian primitives. Fig.~\ref{fig:gaussian_steps} reports the reconstruction performance under different Gaussian optimization steps.
The performance improves rapidly during the early optimization stage and then gradually saturates.
This trend suggests that the progressive optimization efficiently establishes
the shared Gaussian scaffold and refines the diffusion-conditioned response.
After sufficient optimization, additional training steps provide only marginal improvement, indicating that the method reaches a stable subject-specific representation.

\begin{figure}[!t]
\centering

\begin{minipage}[t]{0.48\columnwidth}
    \centering
    \includegraphics[width=\linewidth]
    {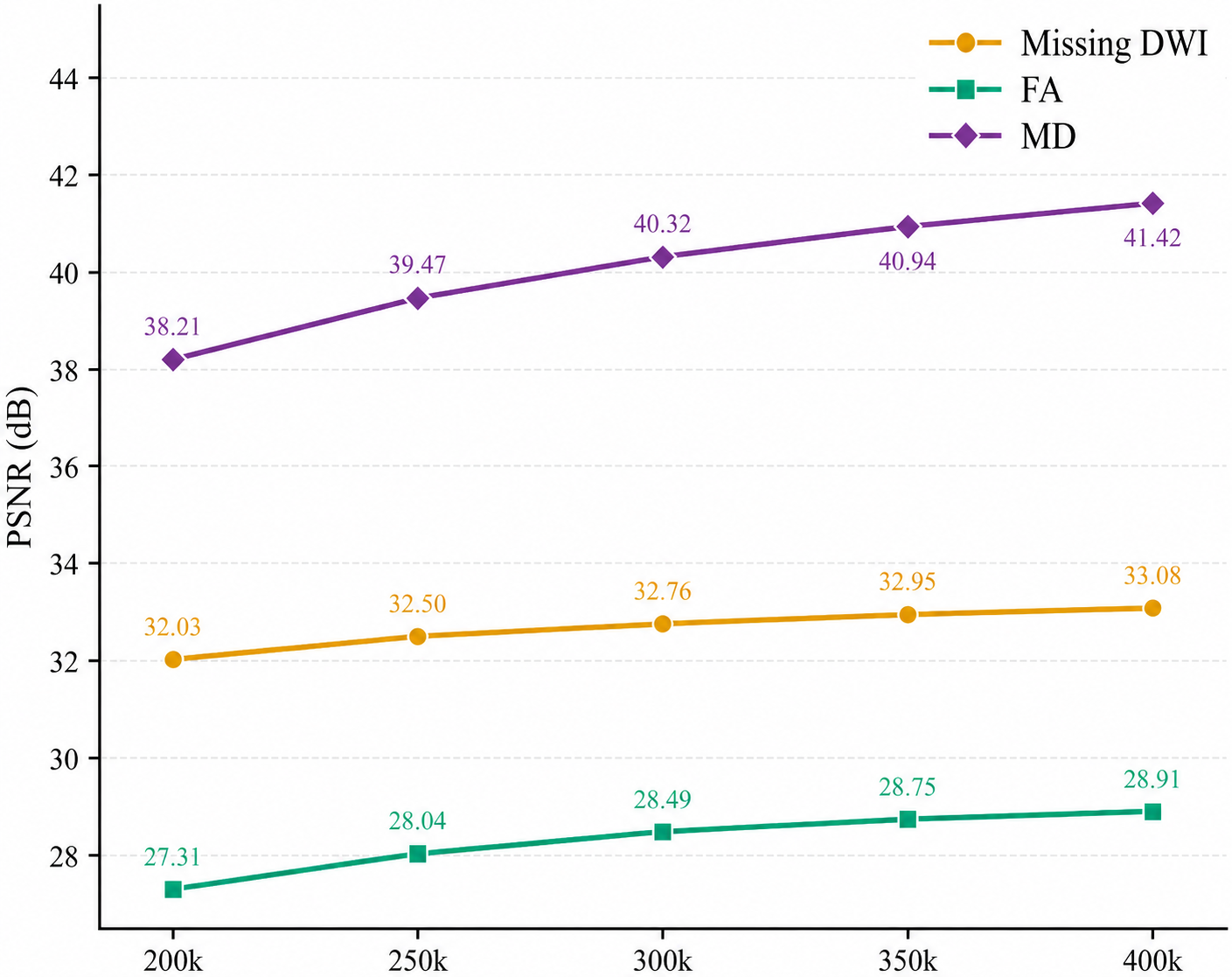}
    \vspace{-1mm}
    \centerline{\small (a)}
\end{minipage}
\hfill
\begin{minipage}[t]{0.48\columnwidth}
    \centering
    \includegraphics[width=\linewidth]
    {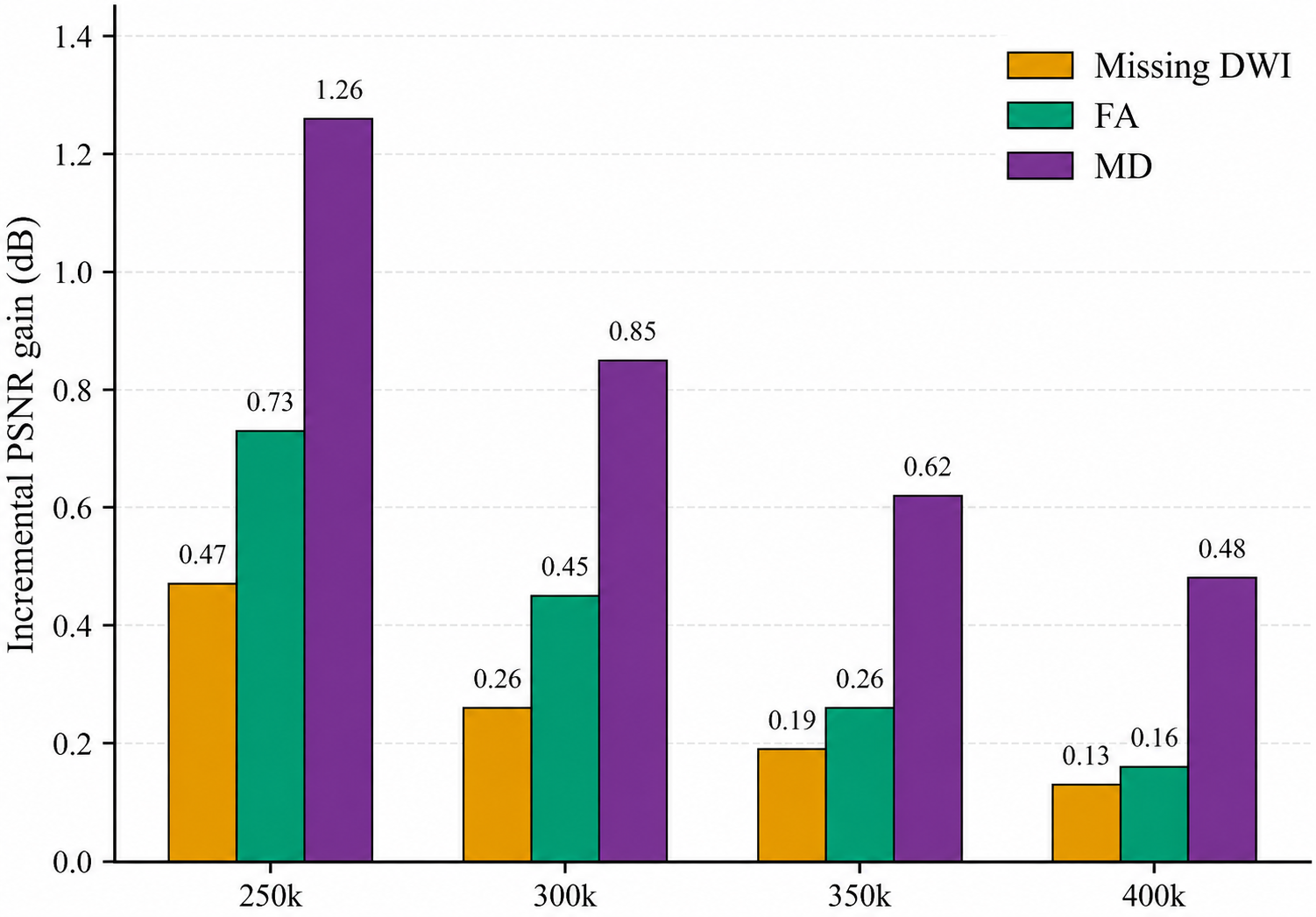}
    \vspace{-1mm}
    \centerline{\small (b)}
\end{minipage}

\vspace{0.5mm}
\caption{
Effect of the Gaussian primitive budget at
$b=1000~\mathrm{s/mm^2}$, $R_k=3$, with 15 observed directions.
(a) DWI, FA, and MD PSNR, (b) incremental PSNR gains.
}
\label{fig:gaussian_primitives}
\end{figure}

Fig.~\ref{fig:gaussian_primitives} evaluates the influence of the final
number of Gaussian primitives after adaptive densification.
As shown in Fig.~\ref{fig:gaussian_primitives}(a), increasing the primitive
budget consistently improves the missing-direction DWI, FA, and MD PSNR,
indicating that additional primitives provide more adequate spatial support
for anatomical and diffusion-dependent structures.
However, the incremental gains in
Fig.~\ref{fig:gaussian_primitives}(b) decrease progressively as the primitive
budget increases.
The improvements from 300k to 350k and from 350k to 400k are smaller than
those obtained at earlier allocation stages, indicating diminishing returns
at larger representation capacities.

\subsection{Ablation Study}
\label{sec:ablation}

We conducted two groups of ablation studies to validate
the main design choices of the proposed framework. The first
group evaluates the primitive-level diffusion response design,
including the tensor branch, SH residual, positive-semidefinite
tensor constraint, and observed-response calibration. The second
group evaluates the optimization and sampling strategy, including
densification, geometry refinement, observed-direction selection,
and the auxiliary zero-filled anchor.

\subsubsection{Ablation on Primitive Response Design}
\label{sec:res_ablation_response}

Table~\ref{tab:ablation_response} summarizes the contribution of the
primitive-level q-response components.
The variant without the tensor branch removes the diffusion-tensor anchor
and retains only an unanchored even-SH response, testing whether angular
flexibility alone can replace the physics-informed main branch.
The variant without the SH residual retains only the tensor-anchored
response, testing whether the tensor component alone is sufficiently
expressive.
The variant without the positive-semidefinite constraint evaluates the
benefit of enforcing physically valid diffusion attenuation.
The variant without observed-response calibration directly optimizes the
continuous tensor--residual response from the undersampled measurements,
testing the benefit of progressive response initialization.

\begin{table}[!t]
\centering
\caption{Ablation of primitive response design at
$b=1000~\mathrm{s/mm^2}$, $R_k=3$, with 15 observed directions.}
\label{tab:ablation_response}

\setlength{\tabcolsep}{5pt}
\renewcommand{\arraystretch}{1.05}

\resizebox{\columnwidth}{!}{%
\begin{tabular}{lccc}
\toprule
Variant
& Missing DWI
& FA
& MD \\
\midrule

\textbf{Full model}
& \textbf{32.58/0.9816}
& \textbf{28.62/0.9646}
& \textbf{40.14/0.9950} \\

w/o tensor branch
& 31.06/0.9731
& 27.32/0.9502
& 37.71/0.9925 \\

w/o SH residual
& 32.40/0.9812
& 28.32/0.9617
& 39.04/0.9945 \\

w/o PSD constraint
& 32.07/0.9800
& 28.41/0.9624
& 39.08/0.9945 \\

w/o response calibration
& 30.85/0.9749
& 27.08/0.9511
& 37.73/0.9935 \\

\bottomrule
\end{tabular}%
}
\end{table}

The full model achieves the best performance across all metrics.
Removing observed-response calibration causes the largest reductions in
missing-direction DWI and FA accuracy, indicating that
observed-direction calibration provides an important initialization for
continuous-response fitting.
Removing the tensor branch produces a pronounced degradation and
the largest decrease in MD PSNR, showing that the physics-informed tensor
anchor benefits held-out-direction synthesis and downstream diffusion
quantification.
Removing the SH residual results in smaller but consistent reductions,
particularly in MD accuracy, indicating that the residual provides
complementary flexibility beyond the tensor response rather than serving as
an independent branch.
Relaxing the positive-semidefinite constraint also reduces DWI and
tensor-derived accuracy, supporting physically valid diffusion attenuation.

\subsubsection{Ablation on Optimization and Sampling Strategy}
\label{sec:res_ablation_optimization}

Table~\ref{tab:ablation_optimization} evaluates adaptive densification,
geometry refinement, observed-direction selection, and the auxiliary
zero-filled anchor while keeping the complete tensor--residual response fixed.
The four corresponding variants respectively remove densification, disable
geometry refinement, replace farthest-point sampling with random direction
selection, and remove the per-direction zero-filled losses during both
response calibration and final refinement.

\begin{table}[!t]
\centering
\caption{Ablation of optimization and sampling strategies at
$b=1000~\mathrm{s/mm^2}$, $R_k=3$, with 15 observed directions.}
\label{tab:ablation_optimization}

\setlength{\tabcolsep}{5pt}
\renewcommand{\arraystretch}{1.05}

\resizebox{\columnwidth}{!}{%
\begin{tabular}{lccc}
\toprule
Variant
& Missing DWI
& FA
& MD \\
\midrule

\textbf{Full model}
& \textbf{32.58/0.9816}
& \textbf{28.62/0.9646}
& \textbf{40.14/0.9950} \\

w/o densification
& 31.19/0.9770
& 26.98/0.9512
& 36.28/0.9912 \\

w/o geometry refinement
& 30.20/0.9716
& 26.58/0.9444
& 34.22/0.9869 \\

Random observed directions
& 31.60/0.9771
& 27.19/0.9519
& 39.00/0.9941 \\

w/o auxiliary ZF anchor
& 32.36/0.9811
& 28.35/0.9623
& 39.05/0.9946 \\

\bottomrule
\end{tabular}%
}
\end{table}

The full model achieves the best performance across all metrics. Disabling geometry refinement causes the largest degradation, reducing
DWI, FA, and MD PSNR by 2.38, 2.04, and 5.92~dB, respectively, which confirms
the importance of jointly refining spatial support and continuous
q-responses. Removing densification also reduces all metrics, particularly MD
PSNR by 3.86~dB, indicating that the initial scaffold provides insufficient
capacity for local diffusion variation. Random direction selection mainly
degrades DWI and FA accuracy, supporting the use of farthest-point sampling
for more uniform q-space coverage. Removing the auxiliary zero-filled anchor has the smallest effect, suggesting that it acts as an auxiliary regularizer,
whereas sampled k-space consistency and scaffold initialization provide the
primary reconstruction constraints.

\section{Discussion}
\label{sec:discussion}

The proposed framework differs from dynamic 4D Gaussian Splatting (4DGS),
which introduces time-dependent deformation or appearance for evolving
scenes. Our representation retains shared spatial 3D Gaussian primitives and
conditions each primitive response on the diffusion-encoding direction. For
a fixed shell, the direction lies on $\mathbb{S}^{2}$ and is modeled through
a continuous tensor--residual response rather than a Euclidean temporal
coordinate. The spatial covariance $\boldsymbol{\Sigma}^{x}_i$ and diffusion
tensor $\mathbf{D}_i$ also belong to different physical domains: the former
defines a primitive's anatomical support, whereas the latter controls signal
attenuation under diffusion gradients. The SH residual complements the
tensor anchor by modeling partial-volume effects and departures from ideal
single-tensor behavior. Its regularized even-SH formulation adds angular
flexibility in the softplus latent domain without replacing the tensor branch
with an unconstrained interpolator. Consistent with
Table~\ref{tab:ablation_response}, the tensor-only and SH-only variants both
underperform the complete tensor--residual response.

The current implementation has several limitations. Experiments use
processed HCP DWI volumes with retrospective single-coil-equivalent in-plane
k-space undersampling rather than raw multi-coil data. A full raw-data
implementation requires coil sensitivity estimation, scanner-specific
trajectories, and noise modeling. The physical response uses one tensor per
Gaussian, which may be limited in complex fiber crossings. Future work can
consider multi-tensor, orientation-distribution, or other
microstructure-aware responses. The spatial Gaussians use diagonal covariance
for robustness and simplicity, while full anisotropic covariance and
optimized CUDA splatting may improve efficiency and spatial adaptivity.
Finally, the framework is scan-specific and self-supervised, and learned
initialization or amortized prediction of Gaussian parameters may reduce
optimization time.

\section{Conclusion}
\label{sec:conclusion}

We presented a physics-informed diffusion-encoding spatial--angular
Gaussian field for self-supervised joint k--q accelerated dMRI
reconstruction.
The proposed method represents subject-specific anatomy using explicit
3D Gaussian primitives and embeds a positive-semidefinite tensor--residual
q-response into each primitive.
By combining differentiable Gaussian splatting, diffusion-tensor physics,
and k-space data consistency, the framework provides a continuous and
interpretable alternative to voxel-grid, image-domain, and black-box
implicit representations.
The method was optimized using only undersampled k-space measurements
from a sparse set of observed directions and supports held-out-direction
synthesis by querying the learned q-response.
Future work will extend the framework to raw multi-coil acquisitions,
multi-shell diffusion encoding, and more expressive microstructural
response models.


\section*{References}
\vspace{-1.3em}
\begin{thebibliography}{99}

\bibitem{r_stejskal1965}
E.~O. Stejskal and J.~E. Tanner,
``Spin diffusion measurements: Spin echoes in the presence of a
time-dependent field gradient,''
\emph{J. Chem. Phys.},
vol.~42, no.~1, pp.~288--292, 1965.

\bibitem{r_basser1994_dti}
P.~J. Basser, J.~Mattiello, and D.~Le Bihan,
``MR diffusion tensor spectroscopy and imaging,''
\emph{Biophys. J.},
vol.~66, no.~1, pp.~259--267, 1994.

\bibitem{r_lebihan2001}
D.~Le Bihan, J.-F.~Mangin, C.~Poupon, C.~A. Clark, S.~Pappata,
N.~Molko, and H.~Chabriat,
``Diffusion tensor imaging: Concepts and applications,''
\emph{J. Magn. Reson. Imaging},
vol.~13, no.~4, pp.~534--546, 2001.

\bibitem{r_basser1996_metrics}
P.~J. Basser and C.~Pierpaoli,
``Microstructural and physiological features of tissues elucidated by
quantitative-diffusion-tensor MRI,''
\emph{J. Magn. Reson. B},
vol.~111, no.~3, pp.~209--219, 1996.

\bibitem{r_kqcs2017}
E.~Schwab, R.~Vidal, and N.~Charon,
``$(k,q)$-compressed sensing for dMRI with joint spatial--angular
sparsity prior,''
in \emph{Computational Diffusion MRI}.
Cham, Switzerland: Springer, 2018, pp.~21--35.

\bibitem{r_qmodel2020}
M.~P. Mani, H.~K. Aggarwal, S.~Ghosh, and M.~Jacob,
``Model-based deep learning for reconstruction of joint k--q
under-sampled high resolution diffusion MRI,''
in \emph{Proc. IEEE Int. Symp. Biomed. Imag. (ISBI)},
2020, pp.~913--916.

\bibitem{r_modl2019}
H.~K. Aggarwal, M.~P. Mani, and M.~Jacob,
``MoDL: Model-based deep learning architecture for inverse problems,''
\emph{IEEE Trans. Med. Imag.},
vol.~38, no.~2, pp.~394--405, Feb.~2019.

\bibitem{r_ssdu2020}
B.~Yaman, S.~A.~H. Hosseini, S.~Moeller, J.~Ellermann,
K.~U\u{g}urbil, and M.~Ak\c{c}akaya,
``Self-supervised learning of physics-guided reconstruction neural
networks without fully sampled reference data,''
\emph{Magn. Reson. Med.},
vol.~84, no.~6, pp.~3172--3191, 2020.

\bibitem{r_qdl2016}
V.~Golkov, A.~Dosovitskiy, J.~I. Sperl, M.~I. Menzel, M.~Czisch,
P.~G. S{\"a}mann, T.~Brox, and D.~Cremers,
``q-Space deep learning: Twelve-fold shorter and model-free diffusion
MRI scans,''
\emph{IEEE Trans. Med. Imag.},
vol.~35, no.~5, pp.~1344--1351, May~2016.

\bibitem{r_msrqdl2021}
Y.~Qin, Y.~Li, Z.~Zhuo, Z.~Liu, Y.~Liu, and C.~Ye,
``Multimodal super-resolved q-space deep learning,''
\emph{Med. Image Anal.},
vol.~71, 2021, Art.~no.~102085.

\bibitem{r_aqdl2025}
F.~Zong, Z.~Zhu, J.~Zhang, X.~Deng, Z.~Li, C.~Ye, and Y.~Liu,
``Attention-based q-space deep learning generalized for accelerated
diffusion magnetic resonance imaging,''
\emph{IEEE J. Biomed. Health Inform.},
vol.~29, no.~2, pp.~1176--1188, Feb.~2025.

\bibitem{r_rcnn2022}
M.~Lyon, P.~Armitage, and M.~A. {\'A}lvarez,
``Angular super-resolution in diffusion MRI with a 3D recurrent
convolutional autoencoder,''
in \emph{Proc. 5th Int. Conf. Medical Imaging with Deep Learning
(MIDL)}, ser. Proc. Mach. Learn. Res.,
vol.~172, pp.~834--846, 2022.

\bibitem{r_pccnn2023}
M.~Lyon, P.~Armitage, and M.~A. {\'A}lvarez,
``Spatio-angular convolutions for super-resolution in diffusion MRI,''
in \emph{Proc. Adv. Neural Inf. Process. Syst.},
vol.~36, pp.~12457--12475, 2023.

\bibitem{r_pgdit2025}
M.~Nan, T.~Xiao, R.~Wu, S.~Yu, Y.~Li, H.~Zheng, and S.~Wang,
``Physics-guided diffusion transformer with spherical harmonic
posterior sampling for high-fidelity angular super-resolution in
diffusion MRI,''
2025, arXiv:2509.07020.

\bibitem{r_siren2020}
V.~Sitzmann, J.~N.~P. Martel, A.~W. Bergman, D.~B. Lindell,
and G.~Wetzstein,
``Implicit neural representations with periodic activation functions,''
in \emph{Proc. Adv. Neural Inf. Process. Syst.},
vol.~33, pp.~7462--7473, 2020.

\bibitem{r_nesh2023}
T.~Hendriks, A.~Vilanova, and M.~Chamberland,
``Neural spherical harmonics for structurally coherent continuous
representation of diffusion MRI signal,''
in \emph{Computational Diffusion MRI},
ser. Lecture Notes in Computer Science, vol.~14328.
Cham, Switzerland: Springer, 2023, pp.~1--12.

\bibitem{r_sarl2025}
R.~Wu, J.~Cheng, C.~Li, J.~Zou, W.~Fan, X.~Ma, H.~Guo,
Y.~Liang, and S.~Wang,
``Spherical harmonics representation learning for high-fidelity and
generalizable super-resolution in diffusion MRI,''
\emph{IEEE Trans. Biomed. Eng.},
vol.~73, no.~4, pp.~1483--1492, Apr.~2026.

\bibitem{r_sainr2026}
Y.~Wu, H.~Rui, F.~Wang, J.~Huang, Z.~Wang, and G.~Yang,
``Self-supervised spatial and zero-shot angular super-resolution by
spatial-angular implicit representation for rotating-view SNR-efficient
diffusion MRI,''
2026, arXiv:2605.02575.

\bibitem{r_3dgs2023}
B.~Kerbl, G.~Kopanas, T.~Leimk{\"u}hler, and G.~Drettakis,
``3D Gaussian Splatting for real-time radiance field rendering,''
\emph{ACM Trans. Graph.},
vol.~42, no.~4, 2023, Art.~no.~139.

\bibitem{r_3dgsmr2025}
T.~Peng, R.~Zha, Z.~Li, X.~Liu, and Q.~Zou,
``Three-dimensional MRI reconstruction with 3D Gaussian
representations: Tackling the undersampling problem,''
\emph{IEEE Trans. Med. Imag.},
vol.~45, no.~5, pp.~1905--1917, May~2026.

\bibitem{r_lustig2007}
M.~Lustig, D.~Donoho, and J.~M. Pauly,
``Sparse MRI: The application of compressed sensing for rapid MR
imaging,''
\emph{Magn. Reson. Med.},
vol.~58, no.~6, pp.~1182--1195, 2007.

\bibitem{r_tuch2004}
D.~S. Tuch,
``Q-ball imaging,''
\emph{Magn. Reson. Med.},
vol.~52, no.~6, pp.~1358--1372, 2004.

\bibitem{r_descoteaux2007}
M.~Descoteaux, E.~Angelino, S.~Fitzgibbons, and R.~Deriche,
``Regularized, fast, and robust analytical Q-ball imaging,''
\emph{Magn. Reson. Med.},
vol.~58, no.~3, pp.~497--510, 2007.

\bibitem{r_sjolund2016}
J.~Sj{\"o}lund, A.~Eklund, E.~{\"O}zarslan, and H.~Knutsson,
``Gaussian process regression can turn non-uniform and undersampled
diffusion MRI data into diffusion spectrum imaging,''
in \emph{Proc. IEEE 14th Int. Symp. Biomed. Imag. (ISBI)},
2017, pp.~778--782.

\bibitem{r_nerf2020}
B.~Mildenhall, P.~P. Srinivasan, M.~Tancik, J.~T. Barron,
R.~Ramamoorthi, and R.~Ng,
``NeRF: Representing scenes as neural radiance fields for view
synthesis,''
in \emph{Proc. Eur. Conf. Comput. Vis. (ECCV)},
2020, pp.~405--421.

\bibitem{r_nerp2021}
L.~Shen, J.~M. Pauly, and L.~Xing,
``NeRP: Implicit neural representation learning with prior embedding
for sparsely sampled image reconstruction,''
\emph{IEEE Trans. Neural Netw. Learn. Syst.},
vol.~35, no.~1, pp.~770--782, Jan.~2024.

\bibitem{r_4dgs2023}
G.~Wu, T.~Yi, J.~Fang, L.~Xie, X.~Zhang, W.~Wei, W.~Liu,
Q.~Tian, and X.~Wang,
``4D Gaussian Splatting for real-time dynamic scene rendering,''
in \emph{Proc. IEEE/CVF Conf. Comput. Vis. Pattern Recognit. (CVPR)},
2024, pp.~20310--20320.

\bibitem{r_r2gaussian2024}
R.~Zha, T.~J. Lin, Y.~Cai, J.~Cao, Y.~Zhang, and H.~Li,
``R$^2$-Gaussian: Rectifying radiative Gaussian Splatting for
tomographic reconstruction,''
in \emph{Proc. Adv. Neural Inf. Process. Syst.},
vol.~37, pp.~44907--44934, 2024.

\bibitem{r_4drgs2024}
Z.~Liu, R.~Zha, H.~Zhao, H.~Li, and Z.~Cui,
``4DRGS: 4D radiative Gaussian Splatting for efficient 3D vessel
reconstruction from sparse-view dynamic DSA images,''
in \emph{Information Processing in Medical Imaging},
ser. Lecture Notes in Computer Science, vol.~15830.
Cham, Switzerland: Springer, 2026, pp.~361--374.

\bibitem{r_4dctgs2025}
Y.~Fu, H.~Zhang, W.~Cai, H.~Xie, L.~Kuo, L.~Cervino,
J.~Moran, X.~Li, and T.~Li,
``Dynamic cone beam CT reconstruction via spatiotemporal Gaussian
neural representation,''
\emph{Med. Phys.},
vol.~52, no.~11, 2025, Art.~no.~e70127.

\bibitem{r_hcp2013}
D.~C. Van Essen \emph{et al.},
``The WU-Minn Human Connectome Project: An overview,''
\emph{NeuroImage},
vol.~80, pp.~62--79, 2013.

\bibitem{r_hcp2016}
M.~F. Glasser \emph{et al.},
``The Human Connectome Project's neuroimaging approach,''
\emph{Nat. Neurosci.},
vol.~19, no.~9, pp.~1175--1187, 2016.



\end{thebibliography}
\end{document}